\documentclass[letterpaper, 10 pt, journal, twoside]{IEEEtran}
\usepackage{graphics} 
\usepackage{epsfig} 
\usepackage{amsmath} 
\usepackage{amssymb}  
\usepackage{color}
\usepackage{xcolor}
\usepackage{cite}
\usepackage{booktabs}
\usepackage{multirow}
\usepackage{hyperref}

\DeclareMathAlphabet{\mathpzc}{T1}{pzc}{m}{it}
\definecolor{blizzardblue}{rgb}{0.67, 0.9, 0.93}
\definecolor{aqua}{rgb}{0.0, 1.0, 1.0}
\definecolor{babyblue}{rgb}{0.54, 0.81, 0.94}

\begin{document}

\title{PTP: Parallelized Tracking and Prediction with Graph Neural Networks and Diversity Sampling}

\author{Xinshuo Weng*, Ye Yuan* and Kris Kitani \vspace{-0.6cm}%
\thanks{Manuscript received: October 15th, 2020; Revised January 13, 2021; Accepted March 1, 2021.}
\thanks{This paper was recommended for publication by Editor Cesar Cadena Lerma upon evaluation of the Associate Editor and Reviewers' comments. This work was supported in part by Qualcomm.} 
\thanks{*The first two authors contributed equally to this work.}
\thanks{All three authors are with Robotics Institute, Carnegie Mellon University. {\tt\footnotesize \{xinshuow, yyuan2, kkitani\}@cs.cmu.edu.}}
\thanks{Digital Object Identifier (DOI): see top of this page.}
}

%
%

\markboth{IEEE Robotics and Automation Letters. Preprint Version. Accepted March, 2021}
{Weng \MakeLowercase{\textit{et al.}}: PTP: Parallelized Tracking and Prediction} 

%



\maketitle

\begin{abstract}
Multi-object tracking (MOT) and trajectory prediction are two critical components in modern 3D perception systems that require accurate modeling of multi-agent interaction. We hypothesize that it is beneficial to unify both tasks under one framework in order to learn a shared feature representation of agent interaction. Furthermore, instead of performing tracking and prediction sequentially which can propagate errors from tracking to prediction, we propose a parallelized framework to mitigate the issue. Also, our parallel track-forecast framework incorporates two additional novel computational units. First, we use a feature interaction technique by introducing Graph Neural Networks (GNNs) to capture the way in which agents interact with one another. The GNN is able to improve discriminative feature learning for MOT association and provide socially-aware contexts for trajectory prediction. Second, we use a diversity sampling function to improve the quality and diversity of our forecasted trajectories. The learned sampling function is trained to efficiently extract a variety of outcomes from a generative trajectory distribution and helps avoid the problem of generating duplicate trajectory samples. We evaluate on KITTI and nuScenes datasets showing that our method with socially-aware feature learning and diversity sampling achieves new state-of-the-art performance on 3D MOT and trajectory prediction. Project website is: \url{http://www.xinshuoweng.com/projects/PTP}.
\end{abstract}


%
\IEEEpeerreviewmaketitle

\section{Introduction}

\IEEEPARstart{T}{RACKING} and trajectory forecasting are critical components in modern 3D perception systems \cite{Wang2018, Weng2020_SPF2}. Historically, 3D multi-object tracking (MOT) \cite{Weng2020_AB3DMOT,Zhang2019,Frossard2018} and trajectory forecasting \cite{alahi2016,Gupta2018,Kosaraju2019,Ivanovic2019,Lee2017,Chandra2019,Deo2018,Yuan2021_AgentFormer,Li2019} have been studied separately. As a result, perception systems often perform 3D MOT and forecasting separately in a cascaded order, where tracking is performed first to obtain trajectories in the past, followed by trajectory forecasting to predict future trajectories. However, this cascaded pipeline with separately trained modules can lead to sub-optimal performance, as information is not shared across two modules during training. Since tracking and forecasting modules are mutually dependent, it would be beneficial to optimize them jointly. For example, a better MOT module can lead to better performance of its downstream forecasting module while a more accurate motion model learned by trajectory forecasting can improve data association in MOT. Our goal is to jointly optimize MOT and forecasting modules and learn a better shared feature representation for both modules.



\begin{figure}[t]
\begin{center}
\includegraphics[trim=0cm 0cm 5cm 0cm, clip=true, width=0.95\linewidth]{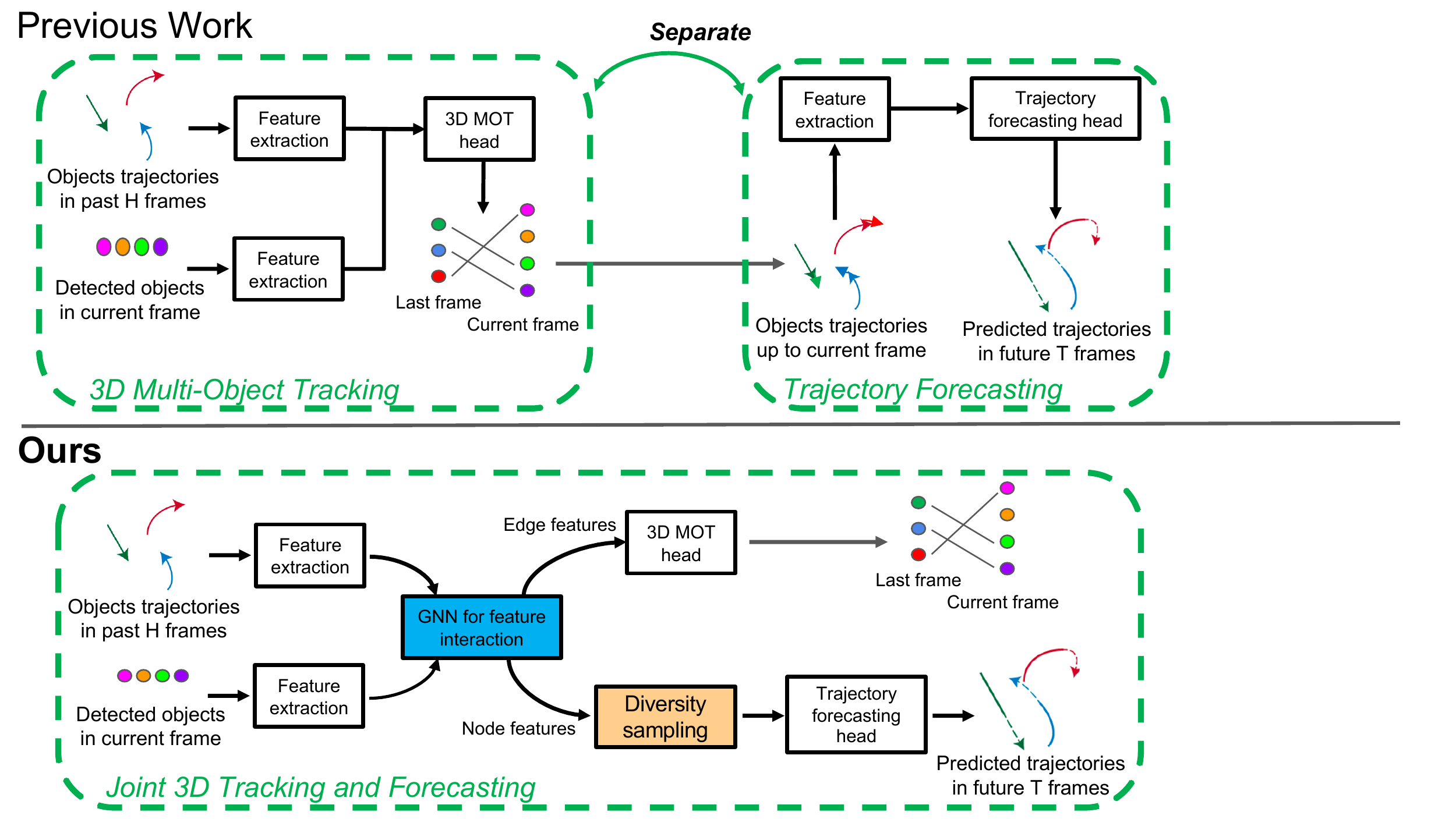}\\
\vspace{-0.35cm}
\caption{\textbf{(Top)} Previous work: 3D MOT and trajectory forecasting performed separately and connected as a sequential process. \textbf{(Bottom)} Ours: Joint and parallelized framework for MOT and forecasting. Two key innovations: (1) \textcolor{blue}{feature interaction using GNNs} to obtain socially-aware features in the presence of multiple agents; (2) \textcolor{orange}{diversity sampling} to improve sample efficiency and produce diverse and accurate trajectory samples.}
\label{fig:teaser}
\vspace{-0.8cm}
\end{center}
\end{figure}


In addition to joint optimization, we also propose to parallelize MOT and forecasting in our framework. Instead of performing two modules in a sequential order as shown in Fig.~\ref{fig:teaser} (top), our framework performs MOT and forecasting in \emph{parallel} as shown in Fig.~\ref{fig:teaser} (bottom). By parallelizing MOT and forecasting heads, i.e., forecasting does not explicitly depend on the MOT results, we can prevent association errors made in MOT from degrading forecasting performance. One might argue that the forecasting module in our parallelized framework cannot utilize association information in the current frame. However, we believe that the forecasting module in our framework can implicitly use association information of the current frame encoded in the shared features. We will show in the experiments that our novel parallelized framework outperforms prior cascaded track-forecast framework.

Modeling interaction for 3D MOT is crucial in the presence of multiple agents but is often overlooked in prior work. Prior work in 3D MOT often extracts the feature of each object \textit{independently}, i.e., each object's feature only depends on the object's own inputs (image crop or location). As a result, there is no interaction between objects. We found that \textit{independent feature extraction leads to inferior discriminative feature learning}, and object dependency is the key to obtaining discriminative features. Intuitively, the features of the same object over two frames should be as similar as possible and the features of two different objects should be as different as possible to avoid confusion during data association. This can only be achieved if object features are obtained in a context-aware process, i.e., modeling interactions between objects.

To model object interaction in 3D MOT, we use a feature interaction mechanism as shown in Fig.~\ref{fig:teaser} (Bottom) by introducing Graph Neural Networks (GNNs). Specifically, we construct a graph with each node being an object in the scene. Then, at each layer of the GNNs, each node can update its feature by aggregating features from other nodes. This node feature aggregation is useful because the resulting object features are no longer isolated and can be adapted according to other objects. We observed in our experiments that, after a few GNN layers, affinity matrix becomes more discriminative than the affinity matrix obtained without interaction. In addition to modeling interaction to improve 3D MOT, interaction modeling can also provide socially-aware context to improve trajectory forecasting \cite{Li2019,Kosaraju2019,Chandra2019_2}. To the best of our knowledge, we are the first to use GNNs to model interaction in a unified framework for both 3D MOT and trajectory forecasting tasks. 

As future trajectories of objects are uncertain and multi-modal due to unobserved factors such as intent, prior work in trajectory forecasting often learns the future trajectory distribution with generative models such as conditional variational autoencoders (CVAEs; \cite{Lee2017}) and conditional generative networks (CGANs; \cite{Gupta2018}). At test time, these methods randomly sample a set of future trajectories from the generative model without considering the correlation between samples. As a result, the samples can be very similar and only cover a limited number of modes, leading to poor sample efficiency. This inefficient sampling strategy is harmful in real-time applications because producing a large number of samples can be computationally expensive and lead to high latency. Moreover, without covering all the modes in the trajectory distribution and considering all possible futures, the perception system cannot plan safely, which is detrimental to safety-critical applications such as autonomous driving.

To improve sample efficiency in trajectory forecasting, we depart from the random sampling in prior work and employ a diversity sampling technique that can generate accurate and diverse trajectory samples from a pretrained CVAE model. The idea is to learn a separate sampling network which maps each object's feature to a set of latent codes. The latent codes are then decoded into trajectory samples. In this way, the produced samples are correlated (unlike random sampling where the samples are independent), which allows us to enforce structural constraints such as diversity onto the samples. Specifically, we use determinantal point processes (DPPs; \cite{kulesza2012determinantal}) to optimize the diversity of the samples. Our contributions are summarized as follows: 
\begin{enumerate}
    \item A parallelized framework for 3D MOT and trajectory forecasting to avoid compounding errors and improve performance of both modules via joint optimization;
    \item A GNNs-based feature interaction mechanism that is the first applied to a unified MOT and forecasting framework to improve socially-aware feature learning;
    \item Introducing diversity sampling to multi-agent trajectory forecasting which can produce more accurate and diverse trajectory samples.
    \vspace{-0.05cm}
\end{enumerate}


\vspace{-0.1cm}
\section{Related Work}
\vspace{-0.1cm}

\noindent\textbf{3D Multi-Object Tracking.}
Recent work tackles 3D MOT in an online fashion using a tracking-by-detection pipeline, where performance is mainly affected by two factors: 3D detection quality and discriminative feature learning. To obtain discriminative features, prior work focuses on feature engineering, among which the motion and appearance features are the most popular ones. \cite{Baser2019,Frossard2018,Hu2019} use Convolutional Neural Networks (CNNs) to extract 2D appearance features. To learn 3D appearance features from point clouds, \cite{Zhang2019} proposes a PointNet-based \cite{Cherabier2017} 3D MOT network. To leverage motion features, filter-based \cite{Simon2019,Weng2020_AB3DMOT} and learning-based methods \cite{Baser2019} have been proposed. Although prior work has achieved impressive performance by feature engineering, they extract feature from each object independently and ignore object interactions. Different from prior work, \cite{Weng2020_GNN3DMOT} is the first introducing GNNs to model interaction in 3D MOT, which significantly improve discriminative feature learning. Different from \cite{Weng2020_GNN3DMOT} which only shows the success of introducing GNNs to 3D MOT, we embed GNNs in a unified 3D MOT and trajectory forecasting framework to improve feature learning for both tasks.

\vspace{1mm}\noindent\textbf{Trajectory Prediction} is to predict a sequence of ground positions of target objects in the future. Prior work mostly investigates the target object of people \cite{Kitani2012,alahi2016,robicquet2016,Gupta2018,Kosaraju2019,Ivanovic2019,yuan2019ego} and vehicles \cite{Lee2017,Rhinehart2019,Rhinehart2018,Chandra2019,Deo2018,Li2019}. As the future is multi-modal, \cite{Gupta2018,Ivanovic2019,Kosaraju2019} use probabilistic models for trajectory prediction. Also, as agent behavior can be influenced by others, \cite{Kosaraju2019,Ivanovic2019,Li2019} introduce GNNs to learn interaction-aware features. However, most prior works study trajectory forecasting separately from 3D MOT, while we consider both forecasting and tracking in a unified framework.

\vspace{1mm}\noindent\textbf{Joint 3D Detection, Tracking and Prediction.} A few prior works attempt joint optimization for different combinations of the three modules. \cite{Hu2019,Frossard2018,Wang2020_GNNDetTrk} jointly optimize detector and tracker. \cite{Zeng2019,Casas2018} achieve joint detection and prediction, while skipping the tracking module. \cite{Luo2018} shows results for detection, tracking and forecasting. However, similar to \cite{Zeng2019,Casas2018}, only detection and prediction are jointly optimized in \cite{Luo2018}, while tracking results are obtained by post-processing. Perhaps concurrent work \cite{Liang2020} is the closest to us which also jointly optimizes detection, tracking and forecasting. However, same as prior work, \cite{Liang2020} processes three modules in a sequential order. To the best of our knowledge, we are the first to propose a parallelized tracking and prediction framework to avoid compounding errors, which also has joint optimization of tracking and prediction modules. 

\vspace{1mm}\noindent\textbf{Graph Neural Networks} was proposed in \cite{Gori2005} to process graph-structured data using neural networks. The primary component of GNNs is node feature aggregation, with which the feature of a node can be updated by interacting with other nodes. Recently, significant success has been achieved by introducing GNNs to applications such as semantic segmentation \cite{Chen2019,Zhang2019_graph}, action recognition \cite{Wang2018_2,Li2019_action,Cheng2019,Zhao2019}, object tracking \cite{Gao2019, Weng2020_GNN3DMOT}. Inspired by prior work, the goal of this work is to introduce existing GNNs techniques to a different practical application -- joint MOT and trajectory forecasting, especially in our novel parallelized MOT and forecasting framework.

\vspace{1mm}\noindent\textbf{Diversity Sampling.} Stemming from the M-Best MAP problem \cite{seroussi1994algorithm}, diverse M-Best solutions \cite{batra2012diverse} and multiple choice learning \cite{lee2016stochastic} are able to produce a diverse ensemble of solutions and models. Also, submodular function maximization \cite{hsiao2018creating} has been used for diverse selection of garments from fashion images. Determinantal point processes (DPPs) \cite{kulesza2012determinantal} are also popular probabilistic models for subset selection due to its ability to measure the global diversity and quality within a set. Prior work has applied DPPs for document and video summarization \cite{gong2014diverse}, object detection~\cite{azadi2017learning}, and grasp clustering~\cite{huang2015we}. Sample diversity has also been an active research topic in generative modeling. A majority of this line of research aims to improve the diversity of the data distribution learned by deep generative models, including works that try to alleviate the mode collapse problem in GANs \cite{che2016mode,arjovsky2017wasserstein,gulrajani2017improved,elfeki2018gdpp} and the posterior collapse problem in VAEs \cite{zhao2017infovae,tolstikhin2017wasserstein,he2019lagging}. Recent work \cite{yuan2019diverse,yuan2020dlow} uses DPPs to improve sample diversity in single-agent trajectory prediction and evaluated on a toy dataset. Different from \cite{yuan2019diverse,yuan2020dlow}, we apply diversity sampling to multi-agent trajectory forecasting and evaluate on real large-scale driving datasets.


\section{Approach\label{sec:approach}}

We aim to achieve 3D MOT and bird's eye view trajectory forecasting in parallel. Let $\mathcal{O} = \{\boldsymbol{o}_1, \ldots, \boldsymbol{o}_M\}$ denote the set of past trajectories of $M$ tracked objects. Each past trajectory $\boldsymbol{o}_i = [\boldsymbol{o}_i^{-H}, \ldots, \boldsymbol{o}_i^{-1}]$ consists of the associated detections of the $i$-th tracked object in the past $H$ frames. The associated detection at frame $t\in \{-H, \ldots, -1\}$ is a tuple $\boldsymbol{o}_i^t = [x, y, z, l, w, h, \theta, I]$, where ($x, y, z$) denotes the object center in 3D space, $(l, w, h)$ denotes the object size, $\theta$ is the heading angle, and $I$ is the assigned ID. Let $\mathcal{D} = \{\boldsymbol{d}_1, \ldots, \boldsymbol{d}_N\}$ denote the set of unassociated detections of $N$ objects in the current frame obtained by a 3D object detector. Each unassociated detection $\boldsymbol{d}_j=[x, y, z, l, w, h, \theta]$ is defined similarly to the past associated detections $\boldsymbol{o}_i^t$ except without the assigned ID $I$. The goal of 3D MOT is to associate the current detection $\boldsymbol{d}_j \in \mathcal{D}$ with the past object trajectory $\boldsymbol{o}_i \in \mathcal{O}$ and assign an ID to $\boldsymbol{d}_j$. For trajectory forecasting, the objective is to predict the future trajectories $\mathcal{F} = \{\boldsymbol{f}_i, \ldots, \boldsymbol{f}_M\}$ for all $M$ tracked objects in the past. Each future trajectory $\boldsymbol{f}_i = [\boldsymbol{f}_i^1, \ldots, \boldsymbol{f}_i^T]$ consists of the $x$ and $z$ positions (i.e., 2D position on the ground in a top-down view) of the $i$-th object in future $T$ frames, i.e., $\boldsymbol{f}_i^t = [x, z]$ where $t\in \{1, \ldots, T\}$. 

The entire network of our method to achieve the parallelized MOT and forecasting is shown in Fig.~\ref{fig:teaser} (bottom), which consists of five modules: (1) a feature extractor to encode the feature for the object trajectories in the past and the detections in the current frame; (2) a feature interaction mechanism using GNNs to update the object features based on the features of other objects; (3) a 3D MOT head that computes the affinity matrix for data association between the tracked objects in the past and detected objects in the current frame; (4) a trajectory forecasting head that learns a CVAE to generate future trajectories based on the GNN features and past trajectories; (5) a diversity sampling that can optimize the diversity of the trajectory samples.

\begin{figure*}[t]
\vspace{0.1cm}
\begin{center}
\includegraphics[trim=0cm 6.5cm 0cm 0cm, clip=true, width=0.95\linewidth]{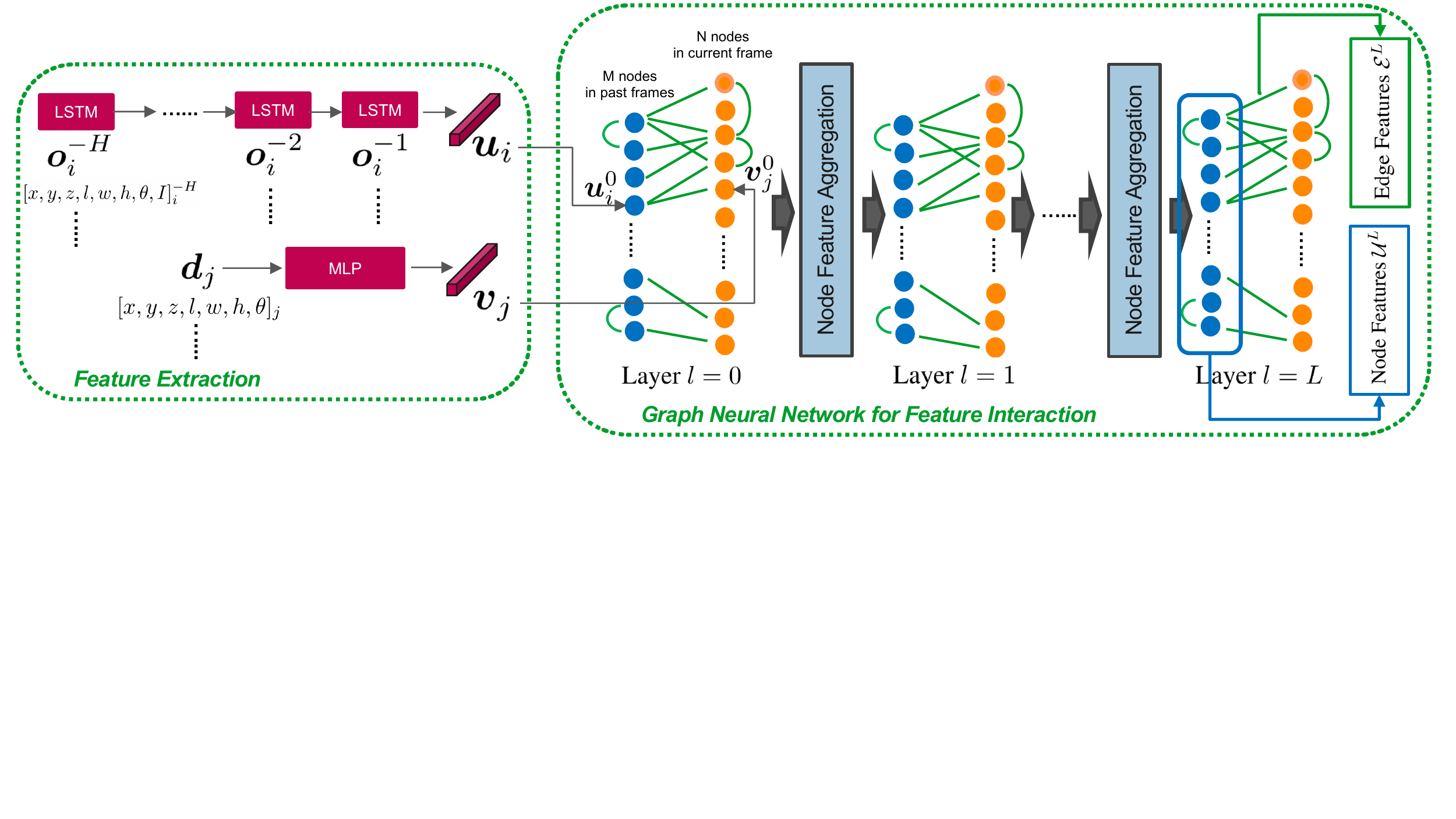}
\vspace{-0.45cm}
\caption{\textbf{(Left)} To leverage the location and motion cues, we extract the feature from object trajectories $\mathcal{O} = \{\boldsymbol{o}_1, \ldots, \boldsymbol{o}_M\}$ in the past using an LSTM model and extract the feature from detections $\mathcal{D} = \{\boldsymbol{d}_1, \ldots, \boldsymbol{d}_N\}$ in the current frame using a MLP. \textbf{(Right)} The GNN-based feature interaction mechanism is used to update object node feature $\mathcal{U}^l = \{\boldsymbol{u}_1^l, \ldots, \boldsymbol{u}_M^l\}$ and $\mathcal{V}^l = \{\boldsymbol{v}_1^l, \ldots, \boldsymbol{v}_N^l\}$ at GNN layer $l$ and iteratively through all GNN layers. At the final layer, we use the node features for tracked objects $\mathcal{U}^L$ for 3D MOT task (see Sec. \ref{sec:mot}), and use the edge features $\mathcal{E}^L$ (computed from $\mathcal{U}^L$ and $\mathcal{V}^L$) for trajectory forecasting task (see Sec. \ref{sec:forecasting} and \ref{sec:diversity}).}
\label{fig:feature}
\vspace{-0.6cm}
\end{center}
\end{figure*}


\vspace{-0.1cm}
\subsection{Feature Extraction}
\vspace{-0.05cm}

To utilize motion and location information from object trajectories in the past and detections in the current frame, we learn feature extractors as shown in Fig.~\ref{fig:feature} (Left). Given the trajectory $\boldsymbol{o}_i = [\boldsymbol{o}_i^{-H}, \ldots, \boldsymbol{o}_i^{-1}]$ for a tracked object $i$, we obtain its feature by applying a two-layer LSTM. The LSTM aims to model the temporal dynamics in the data and outputs a feature $\boldsymbol{u}_i$ with 64 dimensions. For a detected object $j$ in the current frame, we use a 2-layer Multi-Layer Perceptron (MLP) to map the detection $\boldsymbol{d}_j$ to a 64-dimensional feature $\boldsymbol{v}_j$.
Note that the feature extractors for tracked object $\boldsymbol{o}_i$ and detected object $\boldsymbol{d}_j$ are different as $\boldsymbol{o}_i$ and $\boldsymbol{d}_j$ have different time horizon. The obtained features $\boldsymbol{u}_i$ and $\boldsymbol{v}_j$ are then used as initial node features $\boldsymbol{u}_i^0$ and $\boldsymbol{v}_j^0$ at layer 0 of the GNNs (see Sec. \ref{sec:graph}) for feature interaction. 

Note that our feature extractor is shared and optimized for both tracking and forecasting, which is different from prior work that extracts features twice separately in 3D MOT and trajectory forecasting as shown in Fig.~\ref{fig:teaser} (top). As a result, our method reduces system complexity and we will also show in experiments that our parallelized tracking and forecasting framework improves feature learning.


\vspace{-0.1cm}
\subsection{Graph Neural Network for Feature Interaction\label{sec:graph}}
\vspace{-0.05cm}

\noindent\textbf{Graph Construction.} After feature extraction, we have $M$ features $\{\boldsymbol{u}_1^0, \ldots, \boldsymbol{u}_M^0\}$ for tracked objects in the past and $N$ features $\{\boldsymbol{v}_1^0, \ldots, \boldsymbol{v}_N^0\}$ for detected objects in the current frame. We then construct an $L$-layer undirected Graph Neural Network (GNN) where each layer includes nodes of the $M$ tracked objects and $N$ currently detected objects as shown in Fig.~\ref{fig:feature} (right). We use undirected graph because the interaction between objects should be mutual. As the node feature is updated at each layer, let us denote the node features for the tracked objects at layer $l$ as $\mathcal{U}^l= \{\boldsymbol{u}_1^l, \ldots, \boldsymbol{u}_M^l\}$. Similarly, we define the node feature for the currently detected objects at layer $l$ as $\mathcal{V}^l = \{\boldsymbol{v}_1^l, \ldots, \boldsymbol{v}_N^l\}$.

In addition to node feature definition, we also define a set of edges at every layer of the graph to relate node features. To make GNN learning efficient, we restrict edge connections to be sparse, i.e., edges are not defined between every pair of nodes. Specifically, we utilize prior knowledge about social interaction in the presence of multiple agents: interactions primarily happen between objects that are close to each other. Therefore, we construct the edge between two nodes if and only if these two nodes' box centers have distance less than a threshold ($C$ meters) in 3D space. As a result, we have a sparse edge connection as shown in Fig.~\ref{fig:feature} (right). Note that the edge connections are dynamic across time so that GNNs can model interaction in different scenes with varying numbers of objects, though the edge connections are fixed across layers of GNNs at the same time step.

\vspace{1mm}\noindent\textbf{Node Feature Aggregation.} To model feature interaction in GNN, we iteratively update the node features by aggregating features from the neighborhood nodes (i.e., nodes connected by an edge) in each layer. Specifically, we employ the node feature aggregation rule proposed in GraphConv \cite{Morris2019}:
\begin{equation}
   \boldsymbol{u}_i^{l+1} = \sigma_1^l (\boldsymbol{u}_i^l) + \sideset{}{_{j \in \mathpzc{N}(i)}} \sum \sigma_2^l (\boldsymbol{v}_j^l) + \sideset{}{_{g \in \mathpzc{N}(i)}} \sum \sigma_3^l (\boldsymbol{u}_{g}^l),
\end{equation}
\noindent where $\boldsymbol{u}_i^l$ and $\boldsymbol{u}_i^{l+1}$ are the node features for a tracked object $i$ at layer $l$ and $l+1$. $\mathpzc{N}(i)$ denotes a set of neighborhood nodes that are connected to the node $i$ by an edge, and $\boldsymbol{v}_j^l$, $\boldsymbol{u}_g^l$ with $j,g \in \mathpzc{N}(i)$ are the neighborhood node features at layer $l$, where $\boldsymbol{u}_g^l$ is the neighborhood node feature in the past frames and $\boldsymbol{v}_j^l$ is the neighborhood node feature in the current frame. Moreover, $\sigma_1^l$, $\sigma_2^l$, $\sigma_3^l$ are linear layers at layer $l$, whose weights are not shared across layers. Note that a ReLU operator is applied to the node feature after feature aggregation at each layer except for the final layer. Intuitively, the above node aggregation rule means that each node feature is updated by aggregating the transformed features of its own and its connected nodes. In addition to updating the node feature $\boldsymbol{u}_i^l$ for tracked objects, we also update the node feature $\boldsymbol{v}_j^l$ for detected objects:
\begin{equation}
   \boldsymbol{v}_j^{l+1} = \sigma_1^l (\boldsymbol{v}_j^l) + \sideset{}{_{i \in \mathpzc{N}(j)}} \sum \sigma_2^l (\boldsymbol{u}_i^l) + \sideset{}{_{g \in \mathpzc{N}(j)}} \sum \sigma_3^l (\boldsymbol{v}_g^l).
\end{equation}
Based on the above rules, the updated node features for tracked objects $\mathcal{U}^{l+1}$ and for detected objects $\mathcal{V}^{l+1}$ will affect each other through feature interaction in the following layers. After several layers of feature interaction, we use the node features at the final layer $L$ for tracked objects $\mathcal{U}^{L}$ as inputs to our forecasting head (see section \ref{sec:forecasting}). Due to feature interaction, we believe that the final node features $\mathcal{U}^{L}$ have contained enough information from both the trajectories in the past frames and detections in the current frame. 

\vspace{2mm}\noindent\textbf{Edge Feature.} As each entry of the affinity matrix in MOT represents similarity of two objects, it is natural to use edges relating two object nodes to compute the affinity matrix. To learn the similarity, we first define the edge feature between two connected nodes as the difference of their node features:
\vspace{-0.35cm}
\begin{equation}
   \boldsymbol{e}_{ij}^l = \boldsymbol{u}_i^l - \boldsymbol{v}_j^l,
   \vspace{-0.15cm}
\end{equation}
where $\boldsymbol{u}_i^l$ is the node feature of tracked object $i$ in the past frames and $\boldsymbol{v}_j^l$ is the node feature of detected object $j$ in the current frame. The two features are related by an edge feature $\boldsymbol{e}_{ij}^l$ at layer $l$. Note that we only compute the feature for edges relating a tracked object and a detected object as MOT only associates objects across frames (not objects in a same frame). We use the set of edge features $\mathcal{E}^L$ at the final GNN layer as inputs to 3D MOT head for data association. We will show in our experiments that using this simple subtraction between two node features as the edge feature is good enough to achieve S.O.T.A 3D MOT performance.


\vspace{-0.1cm}
\subsection{3D Multi-Object Tracking Head\label{sec:mot}}
\vspace{-0.05cm}

\begin{figure}
\centering
\includegraphics[trim=0cm 10.7cm 15.6cm 0cm, clip=true, width=0.85\linewidth]{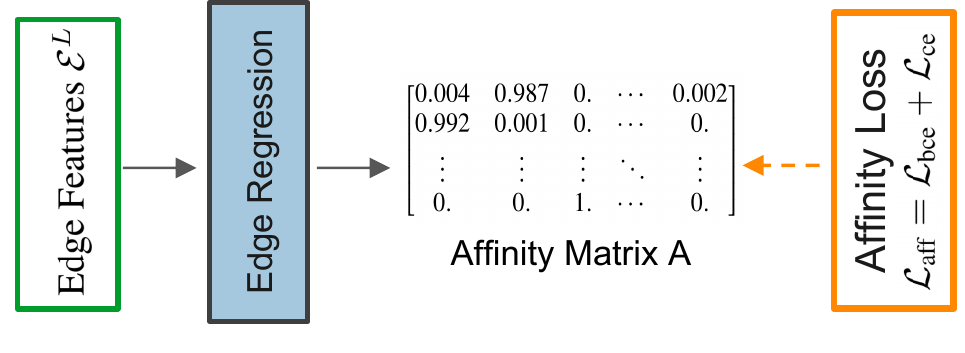}
\vspace{-0.55cm}
\caption{3D multi-object tracking head with an affinity loss.}
\label{fig:edge_regression}
\vspace{-0.7cm}
\end{figure}

To solve 3D MOT, we need to learn an affinity matrix $A$ based on pairwise similarity of the features extracted from $M$ tracked objects in the past and $N$ detected objects in the current frame. As a result, affinity matrix $A$ has a dimension of $M \times N$ where each entry $A_{ij}$ represents the similarity score between the tracked object $i$ and the detected object $j$. 

\vspace{2mm}\noindent\textbf{Edge Regression.} To learn the affinity matrix for data association, we employ an edge regression module as shown in Fig.~\ref{fig:edge_regression}, which consists of a two-layer MLP with a non-linear operator and a Sigmoid layer. To compute each entry in $A_{ij}$, the edge regression module uses an edge feature $\boldsymbol{e}_{ij}^L$ as input and outputs a scalar value between $0$ to $1$ as the pairwise similarity score:
\vspace{-0.15cm}
\begin{equation}
    A_{ij} = \mathrm{Sigmoid} (\sigma_4 (\mathrm{ReLU} (\sigma_3 (\boldsymbol{e}_{ij}^L) ) )),
    \vspace{-0.15cm}
\end{equation}

\noindent where $\sigma_3$ and $\sigma_4$ are two linear layers. As a result, the computed affinity matrix $A$ can be used to associate the objects using the Hungarian algorithm \cite{WKuhn1955} during testing. For tracked and detected objects that cannot be associated, we employ the same birth and death memory as in \cite{Weng2020_AB3DMOT} to create and delete identities. During training, we learn the network parameters by computing an affinity loss between the estimated affinity matrix $A$ and its ground truth (GT).

\vspace{2mm}\noindent\textbf{Affinity Loss.} As shown in Fig.~\ref{fig:edge_regression}, we employ an affinity loss $\mathcal{L}_\mathrm{aff}$ to directly supervise the output $A$ of 3D MOT head. Our affinity loss consists of two individual losses. First, as we know that the GT affinity matrix $A^g$ can only have integer $0$ or $1$ on all the entries, we can formulate the prediction of the affinity matrix as a binary classification problem. Therefore, our first loss is the binary cross entropy (BCE) loss $\mathcal{L}_{\mathrm{bce}}$ that is applied on every entry of $A$:
\begin{equation}
\resizebox{0.9\hsize}{!}{
    $\mathcal{L}_{\mathrm{bce}} = \frac{-1}{MN} \sum_{i=1}^M \sum_{j=1}^N A^g_{ij} \log A_{ij} + (1 - A^g_{ij}) \log (1 - A_{ij}).$
    }
    \vspace{-0.1cm}
\end{equation}
Second, we know that each tracked object $\boldsymbol{o}_i$ can only have either one matched detection $\boldsymbol{d}_j$ or no match at all. In other words, each row and column of the $A^g$ can only be a one-hot vector 
or an all-zero vector. This motivates our second loss. We define the set of rows and columns in $A^g$ that have a one-hot vector as $\mathcal{M}_{oh}$ and $\mathcal{N}_{oh}$, and then apply the cross entropy (CE) loss $\mathcal{L}_{\mathrm{ce}}$ to these rows and columns. As an example, if the $j$th column $A^g_{\cdot j}$ in GT affinity matrix is a one-hot vector, then the loss $\mathcal{L}_{\mathrm{ce}}$ for the $j$th column is defined as:
\vspace{-0.1cm}
\begin{equation}
    \mathcal{L}_{\mathrm{ce}}^{.j} = -\frac{1}{M} \sum_{i=1}^M  A^g_{ij} \log \left( \frac{\exp{A_{ij}}}{\sum_{i=1}^M  \exp{A_{ij}}}  \right).
    \vspace{-0.1cm}
    \label{eq:ce}
\end{equation}

We can summarize the affinity loss $\mathcal{L}_\mathrm{aff}$ for 3D MOT:
\vspace{-0.1cm}
\begin{equation}
    \mathcal{L}_{\mathrm{aff}} = \mathcal{L}_{\mathrm{bce}} + \mathcal{L}_{\mathrm{ce}} = \mathcal{L}_{\mathrm{bce}} + \sum_{i \in \mathcal{M}_{oh}} \mathcal{L}_{\mathrm{ce}}^{i.} + \sum_{j \in \mathcal{N}_{oh}} \mathcal{L}_{\mathrm{ce}}^{.j},
    \vspace{-0.1cm}
\end{equation}
where $\mathcal{L}_{\mathrm{ce}}$ is computed by summing over the Eq. \ref{eq:ce} for all rows and columns with a one-hot vector. Also, we use the same weight of $1$ for two losses $\mathcal{L}_{\mathrm{ce}}$ and $\mathcal{L}_{\mathrm{bce}}$.


\vspace{-0.1cm}
\subsection{Trajectory Forecasting Head\label{sec:forecasting}}
\vspace{-0.05cm}

Our trajectory forecasting head is a conditional generative model $p_\theta(\boldsymbol{f}_i| \boldsymbol{o}_i, \boldsymbol{u}_i^L)$, which learns the distribution of the $i$-th tracked object's future trajectory $\boldsymbol{f}_i$ based on its past trajectory $\boldsymbol{o}_i$ and node feature $\boldsymbol{u}_i^L$ at the last GNN layer. Note that the trajectory forecasting head does not explicitly depend on the MOT association results in the current frame, but instead uses the node feature $\boldsymbol{u}_i^L$ after feature interaction. This design prevents the association error in the current frame made by MOT from deteriorating the forecasting results, while still allowing the forecasting module to exploit the information in the current frame. This is because the node feature $\boldsymbol{u}_i^L$ already encodes object information in the current frame through interaction. As we share the generative model for all tracked objects $\mathcal{O}$, we drop the subscripts and superscripts for ease of notation and denote the generative model as $p_\theta(\boldsymbol{f}| \boldsymbol{o}, \boldsymbol{u})$. We adopt the CVAE~\cite{Lee2017} as our generative model and introduce a latent variable $\boldsymbol{z}$ to model unobserved factors (e.g., agent intentions) and capture the multi-modal distribution of the future trajectory $\boldsymbol{f}$. Based on the CVAE formulation, we introduce a variational lower bound $\mathcal{V}_{lb}(\boldsymbol{f}; \theta, \phi)$ of the log-likelihood function $\log p_\theta(\boldsymbol{f}| \boldsymbol{o}, \boldsymbol{u})$:
\vspace{-0.1cm}
\begin{equation}
\begin{aligned}
\label{eq:cvae}
	\mathcal{V}_{lb}(\boldsymbol{f}; \theta, \phi) = \; & \mathbb{E}_{q_{\phi}(\boldsymbol{z} | \boldsymbol{f}, \boldsymbol{o}, \boldsymbol{u})}\left[\log p_{\theta}(\boldsymbol{f} | \boldsymbol{z}, \boldsymbol{o}, \boldsymbol{u}))\right] \\
	& -\operatorname{KL}\left(q_{\phi}(\boldsymbol{z} | \boldsymbol{f}, \boldsymbol{o}, \boldsymbol{u}) \| p(\boldsymbol{z})\right),
	\vspace{-0.1cm}
\end{aligned}
\end{equation}
where $p(\boldsymbol{z})=\mathcal{N}(\boldsymbol{0}, \boldsymbol{I})$ is a Gaussian latent prior, $q_{\phi}(\boldsymbol{z} | \boldsymbol{f}, \boldsymbol{o}, \boldsymbol{u}) = \mathcal{N}(\boldsymbol{\mu}, \text{Diag}(\boldsymbol{\sigma}^2))$ is an approximated posterior (encoder distribution) and $p_{\theta}(\boldsymbol{f} | \boldsymbol{z}, \boldsymbol{o}, \boldsymbol{u}) = \mathcal{N}(\tilde{\boldsymbol{f}}, \alpha\boldsymbol{I})$ is a conditional likelihood (decoder distribution) with a coefficient $\alpha$. We use two Recurrent Neural Networks (RNNs) as the encoder $F_\phi$ and decoder $G_\theta$ to respectively output the parameters of the encoder and decoder distributions: $(\boldsymbol{\mu}, \boldsymbol{\sigma}) = F_\phi(\boldsymbol{f}, \boldsymbol{o}, \boldsymbol{u})$ and $\tilde{\boldsymbol{f}} = G_\theta(\boldsymbol{z}, \boldsymbol{o}, \boldsymbol{u})$. The detailed architectures for $F_\phi$ and $G_\theta$ are given in the supplementary materials. Based on above formulation, the loss for our trajectory forecasting head is $\mathcal{L}_\text{cvae} = -\mathcal{V}_{lb}$. As we jointly optimize the tracking and forecasting heads as well as the feature extractors and GNNs, we summarize the overall loss of our network as follows:
\vspace{-0.15cm}
\begin{equation}
    \mathcal{L}_{\mathrm{total}} = \mathcal{L}_{\mathrm{aff}} + \mathcal{L}_\text{cvae} = \mathcal{L}_{\mathrm{aff}} -\mathcal{V}_{lb},
    \vspace{-0.15cm}
\end{equation}
Where the weights are $1$ for $\mathcal{L}_{\mathrm{aff}}$ and $\mathcal{L}_\text{cvae}$. Once the CVAE model is learned, we can produce the $i$-th agent's future trajectories $\boldsymbol{f}_i$ by randomly sampling a set of latent codes $\{\boldsymbol{z}_{i1}, \ldots, \boldsymbol{z}_{iK}\}$ from the latent prior and decode them using the decoder $G_\theta$ into future trajectory samples $\{\boldsymbol{f}_{i1}, \ldots, \boldsymbol{f}_{iK}\}$. However, as random sampling can lead to similar samples and low sample efficiency, we introduce a diversity sampling technique into multi-agent trajectory forecasting that can produce diverse and accurate samples, and improve sample efficiency.

\begin{figure}[t]
\begin{center}
\includegraphics[width=\linewidth]{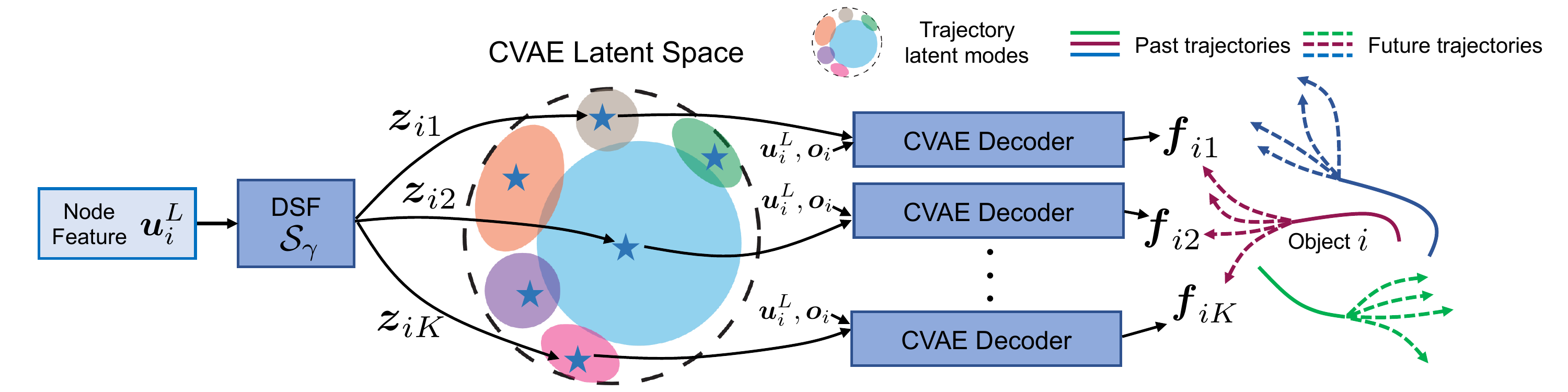}
\vspace{-0.75cm}
\caption{\textbf{Trajectory Forecasting with Diversity Sampling.} To produce diverse trajectory samples from our pretrained CVAE model, we learn a diversity sampling function (DSF) $\mathcal{S}_\gamma$ to map each object's node feature $\boldsymbol{u}_i^L$ to a set of latent codes, which can cover not only the major mode but also other modes in the CVAE latent space. Then, we decode those codes into diverse and accurate future trajectories for object $i$.}
\label{fig:diversiy}
\vspace{-0.7cm}
\end{center}
\end{figure}


\vspace{-0.1cm}
\subsection{Diversity Sampling Technique\label{sec:diversity}}
\vspace{-0.05cm}

To obtain diverse future trajectory samples for multi-agent trajectory forecasting, we introduce the diversity sampling technique. As shown in Fig.~\ref{fig:diversiy}, we use a $\gamma$-parameterized Diversity Sampling Function (DSF) $\mathcal{S}_\gamma$ (a two-layer MLP) that maps the $i$-th object's GNN feature $\boldsymbol{u}_i^L$ to a set of latent codes: $\mathcal{S}_\gamma(\boldsymbol{u}_i^L) = \{\boldsymbol{z}_{i1}, \ldots, \boldsymbol{z}_{iK}\}$. We can then use the CVAE decoder $G_\theta$ to decode the latent codes into a set of future trajectories $\mathcal{Y}_i = \{\boldsymbol{f}_{i1}, \ldots, \boldsymbol{f}_{iK}\}$ for object $i$. In this way, the latent codes and trajectory samples are correlated and can be controlled by the parameters of the DSF $\mathcal{S}_\gamma$. Our goal is to optimize $\mathcal{S}_\gamma$ so that the trajectory samples $\mathcal{Y}_i$ are both diverse and accurate. To model diversity, we construct a DPP kernel $\boldsymbol{L}^i \in \mathbb{R}^{K \times K}$ for each object $i$ based on the diversity and quality of the samples in $\mathcal{Y}_i$:
\vspace{-0.15cm}
\begin{equation}
\boldsymbol{L}^i = \text{Diag}(\boldsymbol{r}^i)\cdot \boldsymbol{S}^i\cdot \text{Diag}(\boldsymbol{r}^i)\,,
\vspace{-0.15cm}
\end{equation}
where the DPP kernel is formed by two components representing sample diversity and quality --- a similarity matrix $\boldsymbol{S}^i \in \mathbb{R}^{K\times K}$ and a quality vector $\boldsymbol{r}^i \in \mathbb{R}^K$:
\vspace{-0.15cm}
\begin{equation}
\begin{aligned}
    \boldsymbol{S}^i_{ab} &= \exp \left(-\omega\|\boldsymbol{f}_{ia} - \boldsymbol{f}_{ib}\|^2 \right),\\
    \boldsymbol{r}^i_k &= \exp\left(\max(-\|\boldsymbol{z}_{ik}\|^2 + R^2, 0 )\right).
\end{aligned}
\vspace{-0.15cm}
\end{equation}
The element $\boldsymbol{S}^i_{ab}$ of the similarity matrix measures similarity between trajectory samples $\boldsymbol{f}_{ia}$ and $\boldsymbol{f}_{ib}$ with a Gaussian kernel where $\omega$ is a scaling factor. Each element $\boldsymbol{r}^i_a$ in the quality vector defines the quality of sample $\boldsymbol{f}_{ia}$ based on how far its latent code $\boldsymbol{z}_{ik}$ is from the origin. If $\boldsymbol{z}_{ik}$ is very far, it means the sample has low likelihood and will be assigned a low quality score. Based on the DPP kernel $\boldsymbol{L}^i$, one can define a diversity loss to measure the diversity and quality within the trajectory samples $\mathcal{Y}_i$:
\vspace{-0.2cm}
\begin{equation}
    \mathcal{L}_\text{dpp}^i = \sum_{n=1}^{N} \frac{\lambda_{n}}{\lambda_{n}+1} =  -\text{tr}\left(\boldsymbol{I} - (\boldsymbol{L}^i + \boldsymbol{I})^{-1}\right),
\vspace{-0.15cm}
\end{equation}
where $\lambda_n$ is the $n$-th eigenvalue of $\boldsymbol{L}^i$, $\text{tr}(\cdot)$ is the trace operator and $\boldsymbol{I}$ is the identity matrix. As the diagonal elements of $\boldsymbol{L}^i$ are all ones, the sum of eigenvalues is fixed: $\sum \lambda_n = \text{tr}(\boldsymbol{L}^i) = K$. The optima of $\mathcal{L}_\text{dpp}^i$ is obtained when all eigenvalues are equal and $\boldsymbol{L}^i$ becomes an identity matrix, thus making $\boldsymbol{S}^i_{ab} = 0$ $(a\neq b)$ and $\boldsymbol{r}^i_k = 1$. This means the distance between trajectory samples is large and each sample has high likelihood. However, this optima is seldom obtained due to the trade-off between diversity and quality, i.e., samples far way from others often have low likelihood. Besides the diversity loss, we further introduce a reconstruction loss to encourage the set of trajectory samples $\mathcal{Y}_i$ to cover the ground-truth future trajectories $\hat{\boldsymbol{f}}$:
\vspace{-0.15cm}
\begin{equation}
    \mathcal{L}_\text{recon}^i = \min_k \|\boldsymbol{f}_{ik} - \hat{\boldsymbol{f}}\|^2\,.
    \vspace{-0.15cm}
\end{equation}
To learn DSF, we freeze the parameters of other components (feature extractors, GNNs, and CVAE) and only optimize the parameters~$\gamma$ of the DSF with the following loss:
\vspace{-0.2cm}
\begin{equation}
    \mathcal{L}_\text{dsf} = \frac{1}{M}\sum_{i=1}^M \mathcal{L}_\text{dpp}^i +  \mathcal{L}_\text{recon}^i \,.
\vspace{-0.2cm}
\end{equation}


\vspace{-0.1cm}
\section{Experiments}
\vspace{-0.05cm}

\subsection{Datasets}
\vspace{-0.05cm}

We evaluate on standard driving datasets: KITTI \cite{Geiger2012} and nuScenes \cite{Caesar2019}, which provide LiDAR point cloud and 3D bounding box trajectories. We do not evaluate on 2D MOT datasets such as MOTChallenges \cite{Milan2016} as they do not provide LiDAR data or 3D bounding box ground truth and are thus not directly applicable to our method. To compare against prior state-of-the-art methods for 3D MOT alone or for trajectory forecasting only, we evaluate two modules separately. For KITTI, same as prior work, we report results on the car subset for comparison. For nuScenes, we evaluate our 3D MOT and trajectory forecasting for all categories (car, pedestrian, bicycle, motorcycle, truck, bus and trailer) and final performance is the mean over all categories.


\vspace{-0.2cm}
\subsection{Evaluating 3D Multi-Object Tracking}

\noindent\textbf{Evaluation Metrics.} We use standard CLEAR metrics (including MOTA, MOTP, IDS) and new sAMOTA, AMOTA and AMOTP metrics \cite{Weng2020_AB3DMOT}. Since we are evaluating 3D MOT methods, all above metrics need to be defined in 3D space using the criteria of 3D IoU or 3D distance. However, KITTI dataset only supports 2D MOT evaluation, i.e., metrics defined in 2D space for evaluating image-based MOT methods. Therefore, instead of using KITTI 2D MOT evaluation, we use 3D MOT evaluation code provided by \cite{Weng2020_AB3DMOT}. 

\vspace{2mm}\noindent\textbf{Baselines.} We compare against recent 3D MOT systems such as FANTrack \cite{Baser2019}, mmMOT \cite{Zhang2019}, AB3DMOT \cite{Weng2020_AB3DMOT} and GNN3DMOT \cite{Weng2020_GNN3DMOT}. To achieve fair comparison, we use the same 3D detections obtained by PointRCNN \cite{Shi2019} on KITTI and by Megvii \cite{Zhu2019} on nuScenes for all methods. Also, for some baselines \cite{Hu2019,Baser2019,Weng2020_GNN3DMOT} that require 2D detections as inputs, we use the 2D projection of the 3D detections. 

\begin{figure}[t]
\begin{center}
\includegraphics[trim=0cm 0cm 0cm 0cm, clip=true, width=0.49\linewidth]{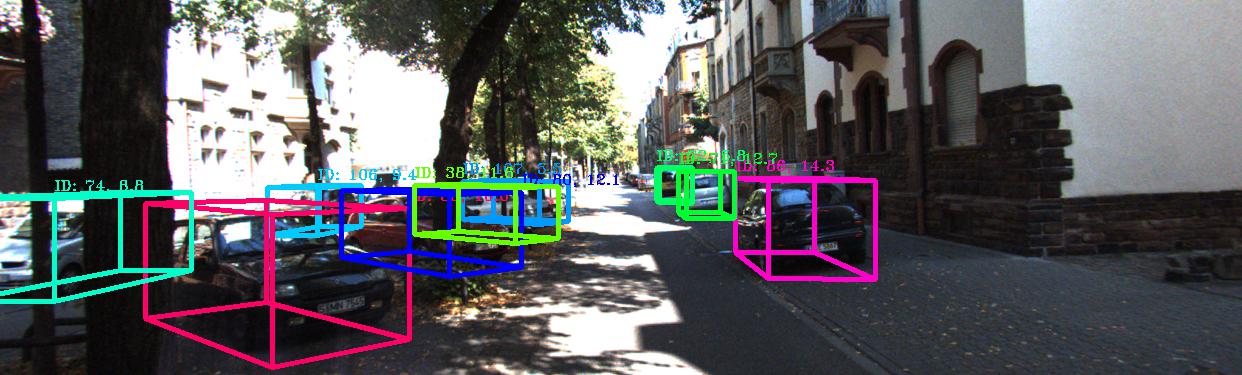}
\includegraphics[trim=0cm 0cm 0cm 0cm, clip=true, width=0.49\linewidth]{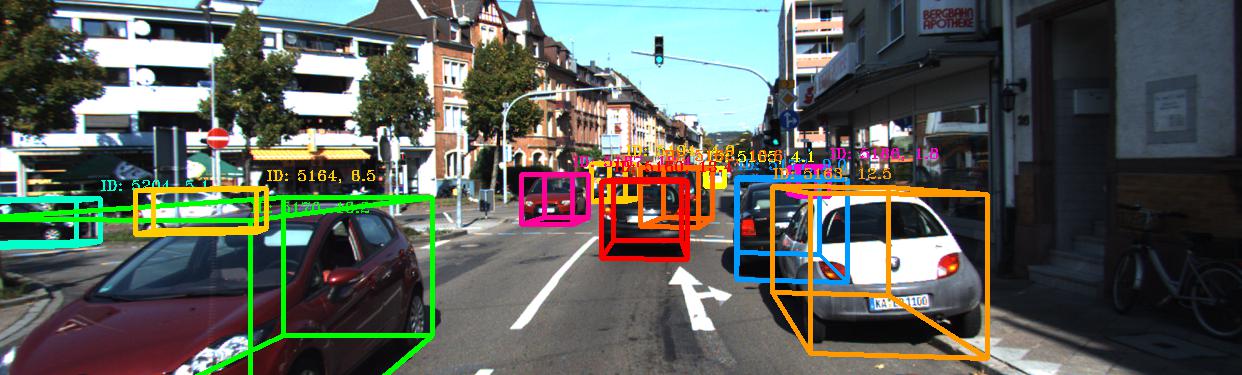}
\includegraphics[trim=0cm 0cm 0cm 0cm, clip=true, width=0.49\linewidth]{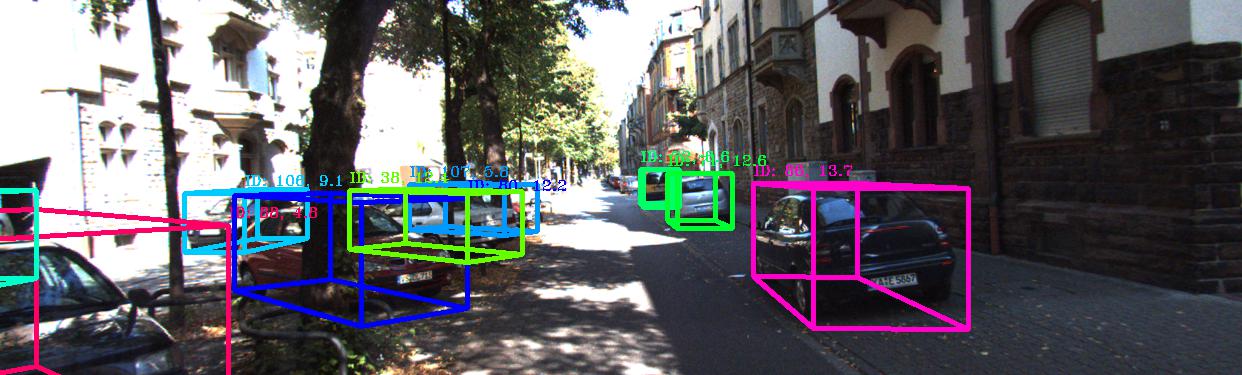}
\includegraphics[trim=0cm 0cm 0cm 0cm, clip=true, width=0.49\linewidth]{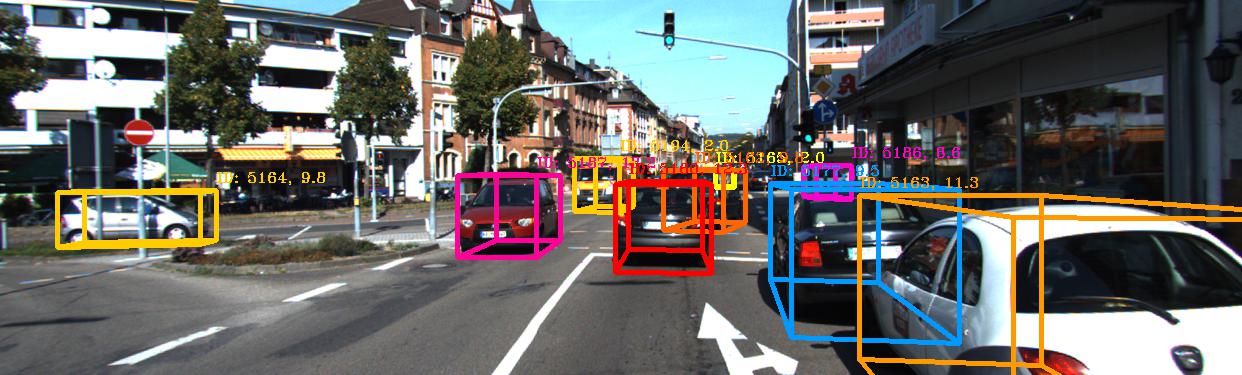}
\includegraphics[trim=0cm 0cm 0cm 0cm, clip=true, width=0.49\linewidth]{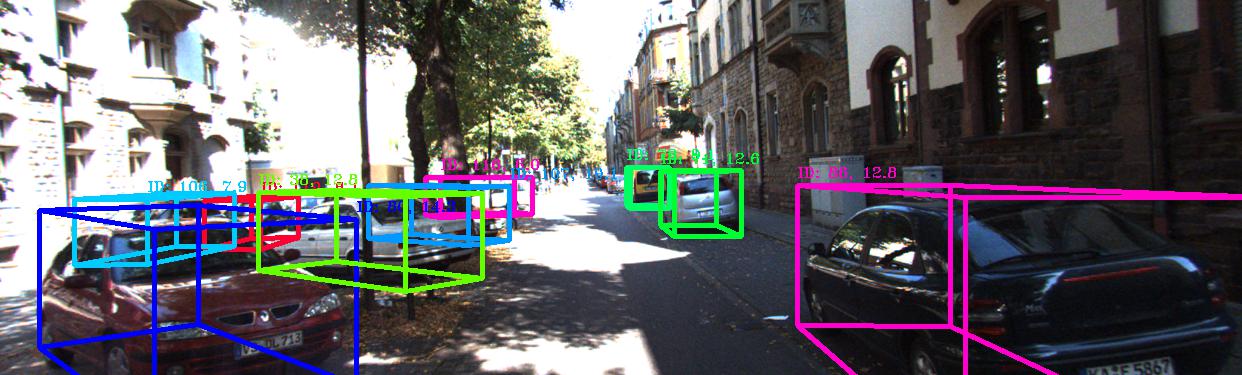}
\includegraphics[trim=0cm 0cm 0cm 0cm, clip=true, width=0.49\linewidth]{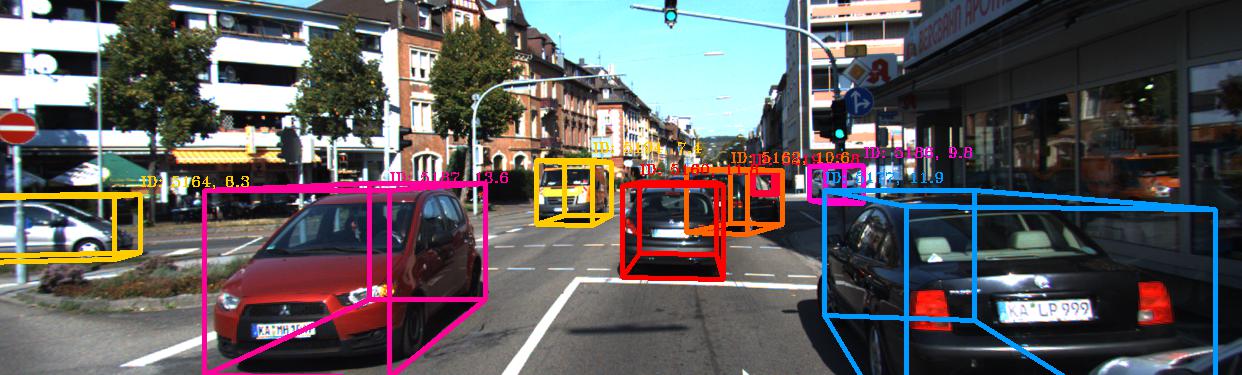}
\end{center}
\vspace{-0.55cm}
\caption{3D MOT results on sequence 0 (left) and 11 (right) of the KITTI test set. We show three frames (with an interval of 5 frames) for each sequence.}
\label{fig:3dmot_qua}
\vspace{-0.25cm}
\end{figure}

\vspace{2mm}\noindent\textbf{Results.} We summarize the 3D MOT results on KITTI and nuScenes datasets in Table \ref{tab:3dmot_quan}. Our method consistently outperforms baselines in sAMOTA, AMOTA and MOTA, which are the primary metrics for ranking MOT methods. We hypothesize that this is because our method leveraging GNN obtains more discriminative features to avoid confusion in MOT association while all 3D MOT baselines ignore the interaction between objects. Moreover, joint optimization of the tracking and forecasting modules in parallel also helps. We will justify both hypotheses in the ablation study. We show qualitative results of our method on the KITTI dataset in Fig. \ref{fig:3dmot_qua}, demonstrating reliable 3D MOT performance.

\begin{table}[t!]
\caption{3D MOT Evaluation on the KITTI and nuScenes datasets.}
\vspace{-0.3cm}
\centering
\resizebox{\linewidth}{!}{
\begin{tabular}{@{}llrrrrrrr@{}}
\toprule
Datasets & Methods  &  \textbf{sAMOTA}(\%)$\uparrow$ & AMOTA(\%)$\uparrow$ &  AMOTP(\%)$\uparrow$ & MOTA(\%)$\uparrow$ &  MOTP(\%)$\uparrow$ & IDS$\downarrow$ \\
\midrule
\multirow{5}{*}{KITTI} 
& mmMOT~\cite{Zhang2019}            & 70.61 & 33.08 & 72.45 & 74.07 & 78.16 & 10\\ 
& FANTrack~\cite{Baser2019}         & 82.97 & 40.03 & 75.01 & 74.30 & 75.24 & 35\\
& AB3DMOT\cite{Weng2020_AB3DMOT}    & 93.28 & 45.43 & 77.41 & 86.24 & 78.43 & \textbf{0} \\
& GNN3DMOT\cite{Weng2020_GNN3DMOT}  & 93.68 & 45.27 & \textbf{78.10} & 84.70 & \textbf{79.03} & \textbf{0} \\
& \textbf{Ours (PTP)} & \textbf{94.41} & \textbf{46.15} & 76.83 & \textbf{86.89} & 78.32 & 3  \\
\midrule
\multirow{5}{*}{nuScenes} 
& FANTrack~\cite{Baser2019}         & 19.64 & 2.36 & 22.92 & 18.60 & 39.82 & 1593 \\
& mmMOT~\cite{Zhang2019}            & 23.93 & 2.11 & 21.28 & 19.82 & 40.93 & 572 \\ 
& GNN3DMOT\cite{Weng2020_GNN3DMOT}  & 29.84 & 6.21 & 24.02 & 23.53 & 46.91 & \textbf{401} \\
& AB3DMOT\cite{Weng2020_AB3DMOT}    & 39.90 & 8.94 & 29.67 & 31.40 & 57.54 & 751 \\
& \textbf{Ours (PTP)} & \textbf{42.36} & \textbf{9.84} & \textbf{33.48} & \textbf{32.06} & \textbf{63.61} & 809 \\
\bottomrule
\end{tabular}}
\vspace{-0.35cm}
\label{tab:3dmot_quan}
\end{table}

\begin{table}[t!]
	\begin{minipage}{0.495\linewidth}
	    \vspace{-0.2cm}
		\caption{Effect of trajectory forecasting on 3D MOT.}
		\vspace{-0.3cm}
		\label{tab:mot_tf}
		\centering
        \resizebox{\textwidth}{!}{
        \begin{tabular}{@{}lrr@{}}
        \toprule
        Metrics \ \ \ & \ \ w/o forecasting & \ \ w/ forecasting \\ 
        \midrule
        \textbf{sAMOTA}(\%)$\uparrow$ & 91.31 & \textbf{94.41} \\ 
        AMOTA(\%)$\uparrow$           & 43.68 & \textbf{46.15} \\ 
        AMOTP(\%)$\uparrow$           & \textbf{76.94} & 76.83 \\ 
        MOTA(\%)$\uparrow$            & 83.51 & \textbf{86.89} \\ 
        MOTP(\%)$\uparrow$            & 78.11 & \textbf{78.32} \\ 
        IDS$\downarrow$               & 5 & \textbf{3}\\ 
        \bottomrule
        \end{tabular}}
	\end{minipage}
	\begin{minipage}{0.495\linewidth}
		\centering
        \includegraphics[trim=0.3cm 0cm 0.3cm 0.3cm, clip=true, width=0.9\linewidth]{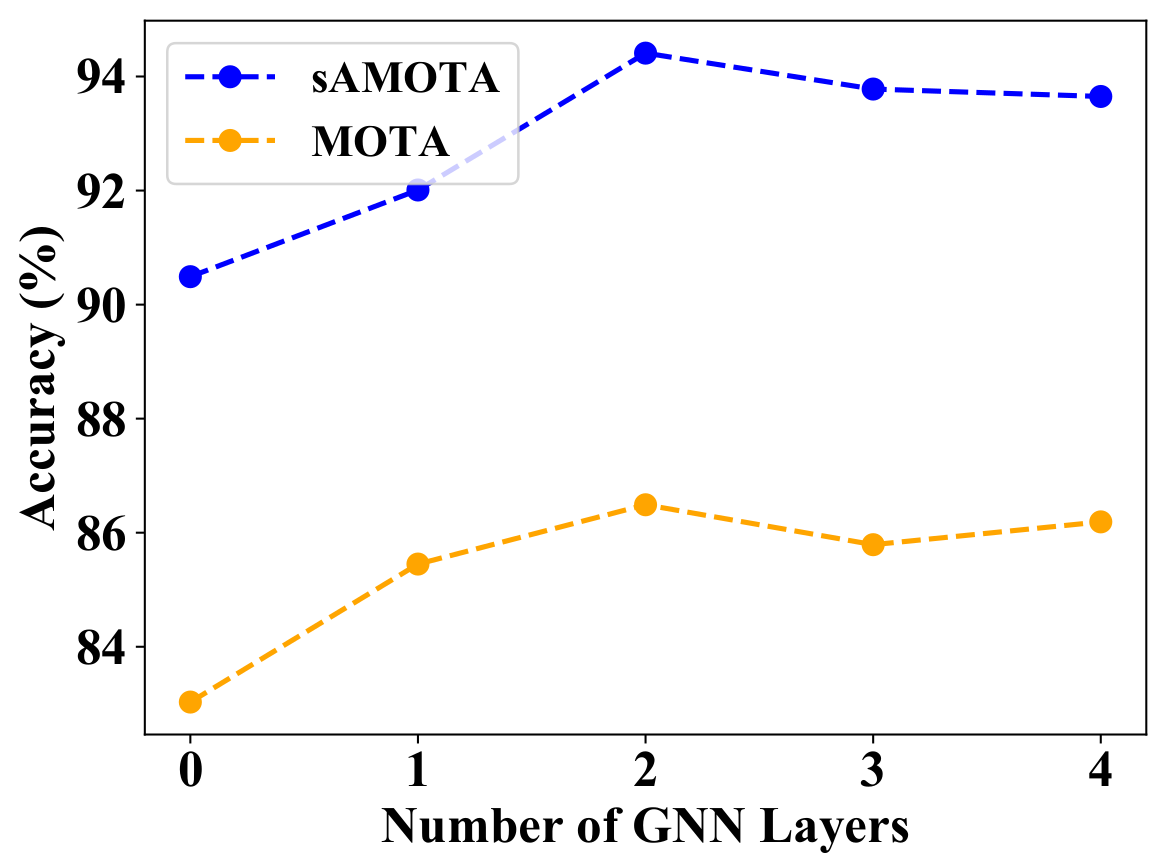}
        \label{fig:mot_gnn}
	\end{minipage}
	\vspace{-0.75cm}
\end{table}

\vspace{2mm}\noindent\textbf{Ablation Study.} We first verify if joint optimization of the 3D MOT and forecasting improves 3D MOT performance. In Table \ref{tab:mot_tf}, when we train MOT and trajectory forecasting heads together on the KITTI dataset, performance is higher in most metrics compared to the model without forecasting. This proves that parallelizing two modules is beneficial to 3D MOT. Second, as shown in the figure next to Table \ref{tab:mot_tf}, on the KITTI dataset we validate the effect of the number of GNN layers on 3D MOT. We can see that, performance is significantly increased when using two layers of GNN, and then starts to converge with more layers. As a result, we use two GNN layers for shared feature learning in our method. 


\begin{table}[t]
\begin{center}
\caption{Trajectory forecasting evaluation on KITTI and nuScenes.}
\vspace{-0.6cm}
\resizebox{\linewidth}{!}{
\begin{tabular}{@{}llrrrrrrr@{}}
\toprule
Datasets & Metrics & Conv-Social \cite{Deo2018} & Social-GAN \cite{Gupta2018} & TraPHic \cite{Chandra2019} & Graph-LSTM \cite{Chandra2019_2} & \textbf{Ours (PTP)} \\ 
\midrule

\multirow{4}{*}{KITTI-1.0s} & ADE$\downarrow$ & 0.607 & 0.586 & 0.542 & 0.478 & \textbf{0.471}\\
                            & FDE$\downarrow$ & 0.948 & 1.167 & 0.839 & 0.800 & \textbf{0.763}\\ 
                            & ASD$\uparrow$   & 1.785 & 0.495 & 1.787 & 1.070 & \textbf{2.351}\\ 
                            & FSD$\uparrow$   & 1.987 & 0.844 & 1.988 & 1.836 & \textbf{4.071}\\ 
\midrule

\multirow{4}{*}{KITTI-3.0s} & ADE$\downarrow$ & 2.362 & 2.340 & 2.279 & 1.994 & \textbf{1.319}\\
                            & FDE$\downarrow$ & 3.916 & 4.102 & 3.780 & 3.351 & \textbf{2.299}\\ 
                            & ASD$\uparrow$   & 2.436 & 1.351 & 2.434 & 2.745 & \textbf{5.843}\\ 
                            & FSD$\uparrow$   & 2.973 & 2.066 & 2.973 & 4.582 & \textbf{10.123}\\ 
\midrule

\multirow{4}{*}{nuScenes-1.0s} & ADE$\downarrow$ & 0.674 & 0.483 & 0.571 & 0.509 & \textbf{0.378}\\ 
                               & FDE$\downarrow$ & 0.784 & 0.586 & 0.640 & 0.618 & \textbf{0.490}\\ 
                               & ASD$\uparrow$   & 2.101 & 1.005 & 2.102 & 1.122 & \textbf{5.665}\\ 
                               & FSD$\uparrow$   & 2.430 & 1.475 & 2.432 & 1.603 & \textbf{7.826}\\ 
\midrule

\multirow{4}{*}{nuScenes-3.0s} & ADE$\downarrow$ & 1.989 & 1.794 & 1.827 & 1.646 & \textbf{1.017}\\ 
                               & FDE$\downarrow$ & 3.015 & 2.850 & 2.760 & 2.445 & \textbf{1.527}\\ 
                               & ASD$\uparrow$   & 2.799 & 1.945 & 2.803 & 2.742 & \textbf{8.323}\\ 
                               & FSD$\uparrow$   & 4.174 & 3.610 & 4.184 & 4.970 & \textbf{15.787}\\ 
\bottomrule
\end{tabular}
\label{tab:trajectory_quan} 
}
\end{center}
\vspace{-0.55cm}
\end{table}

\begin{figure}[t]
\begin{center}
\includegraphics[trim=0cm 0cm 0cm 0cm, clip=true, width=0.49\linewidth]{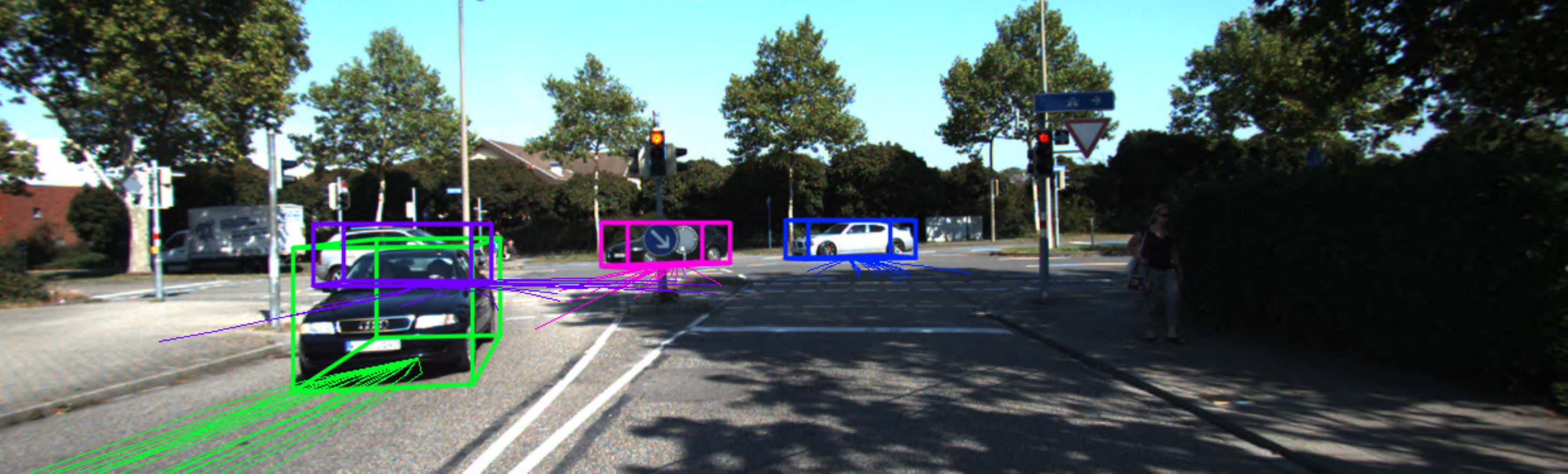}
\includegraphics[trim=0cm 0cm 0cm 0cm, clip=true, width=0.49\linewidth]{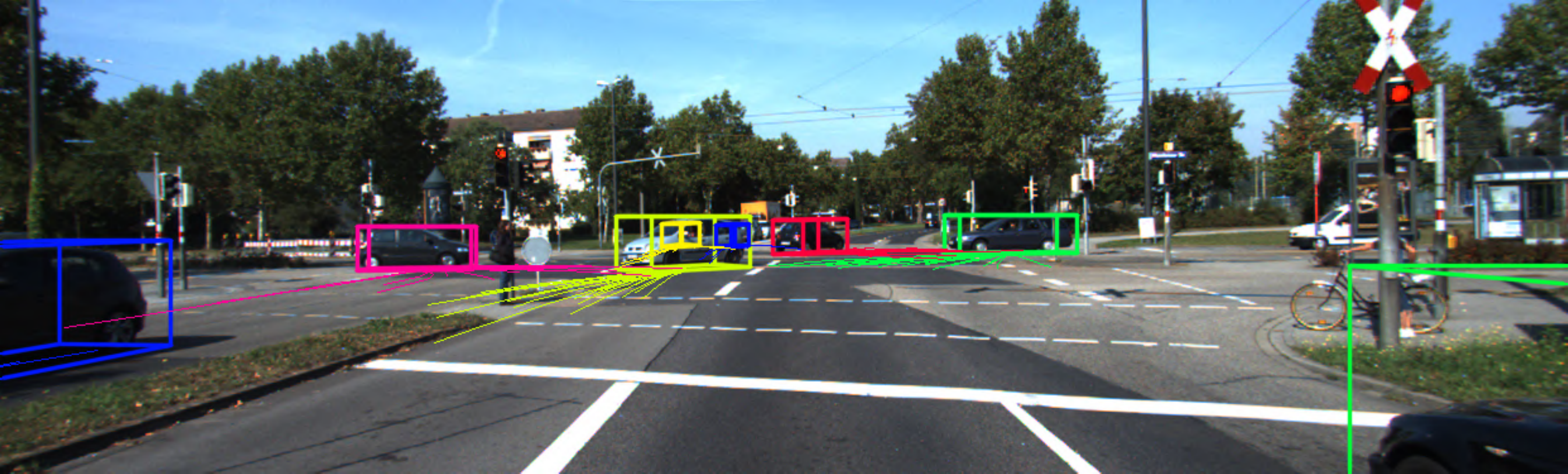}
\includegraphics[trim=0cm 0cm 0cm 0cm, clip=true, width=0.49\linewidth]{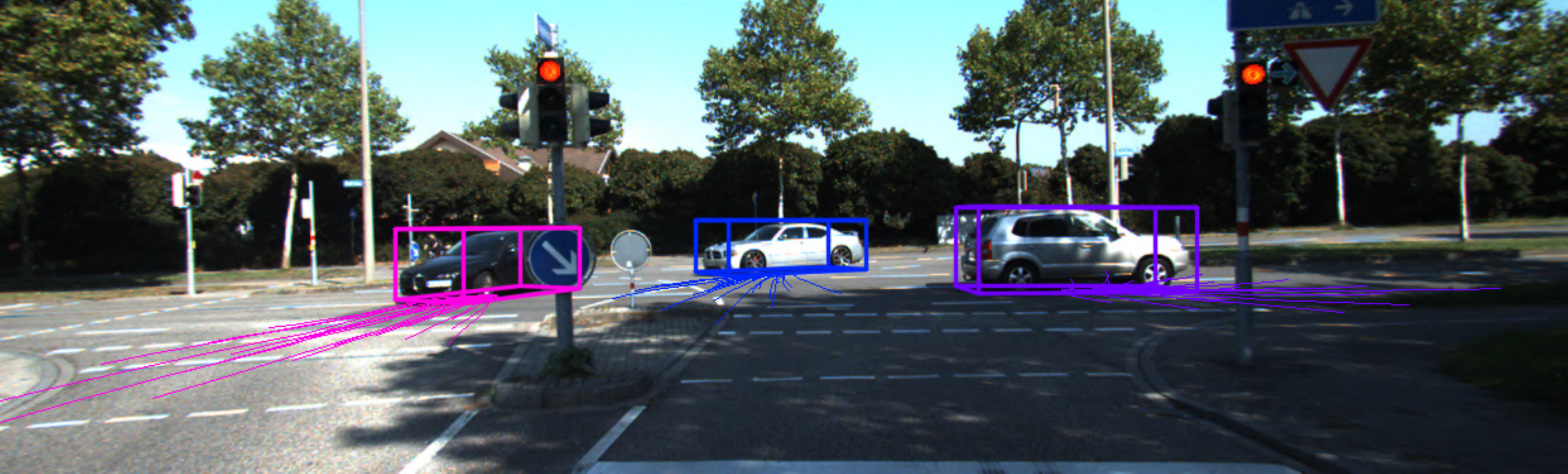}
\includegraphics[trim=0cm 0cm 0cm 0cm, clip=true, width=0.49\linewidth]{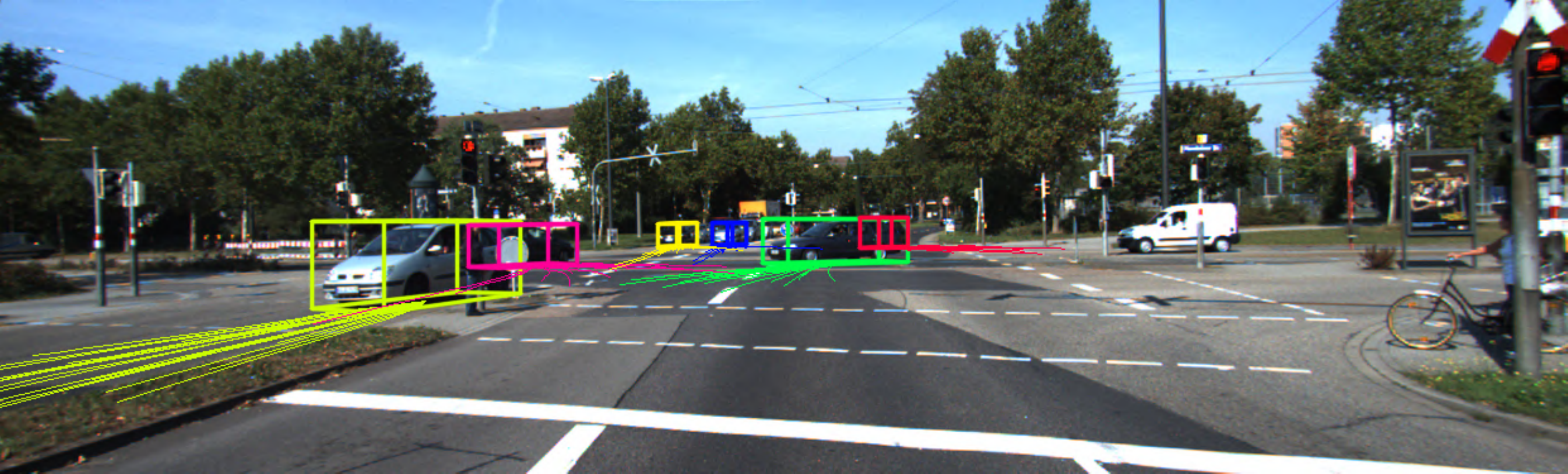}
\includegraphics[trim=0cm 0cm 0cm 0cm, clip=true, width=0.49\linewidth]{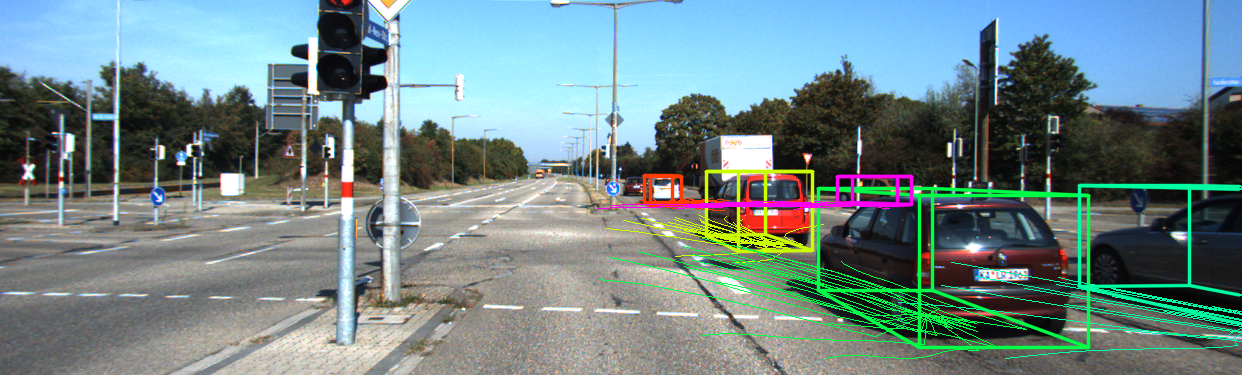}
\includegraphics[trim=0cm 0cm 0cm 0cm, clip=true, width=0.49\linewidth]{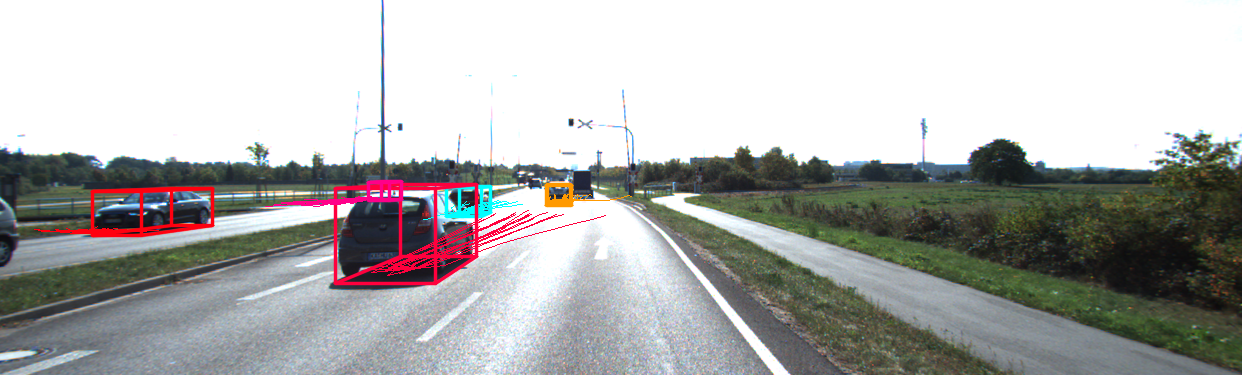}
\includegraphics[trim=0cm 0cm 0cm 0cm, clip=true, width=0.49\linewidth]{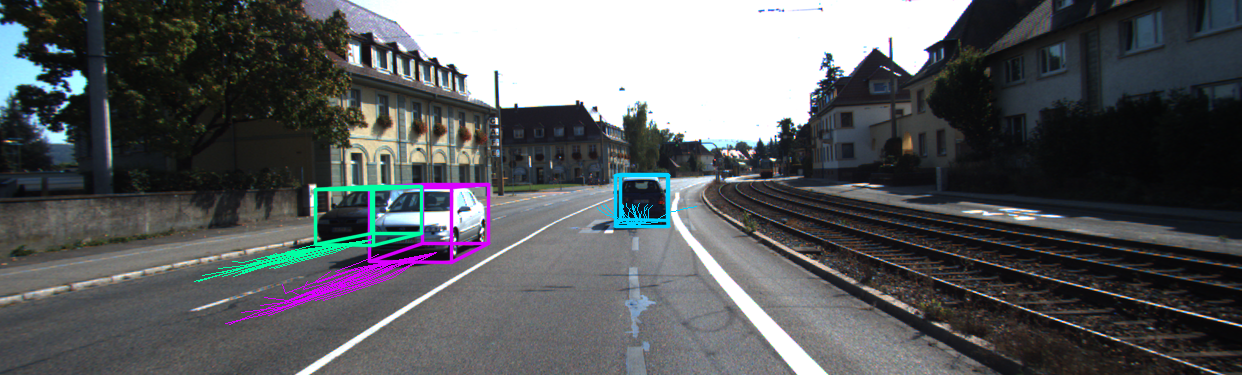}
\includegraphics[trim=0cm 0cm 0cm 0cm, clip=true, width=0.49\linewidth]{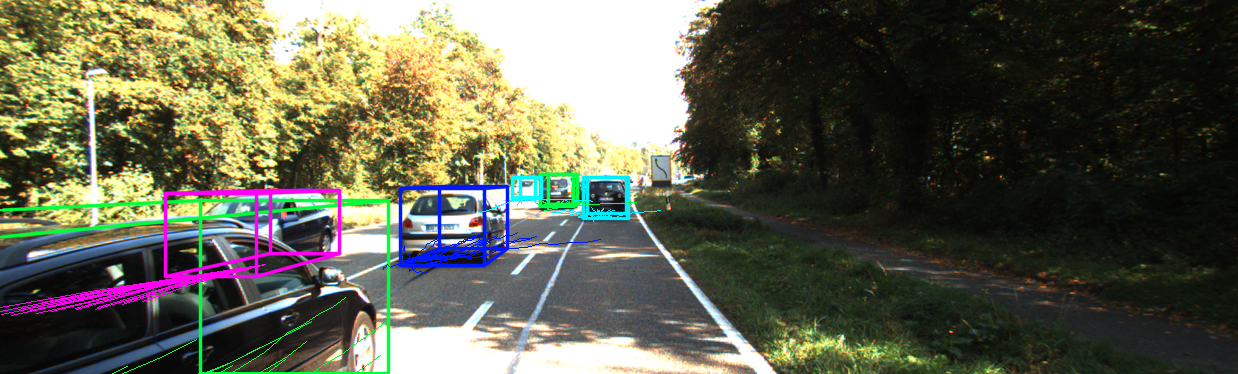}
\vspace{-0.75cm}
\caption{Trajectory forecasting visualization on the KITTI dataset.}
\end{center}
\label{fig:trajectory_qua}
\vspace{-0.8cm}
\end{figure}

\begin{table}[t!]
    \begin{minipage}{0.53\linewidth}
    	\caption{Effect of 3D MOT and diversity sampling on forecasting.}
    	\vspace{-0.3cm}
    	\label{tab:trajectory_ablation}
    	\centering
        \resizebox{\textwidth}{!}{
        \begin{tabular}{@{}llrrrr@{}}
        \toprule
        Datasets  & Metrics  &  w/o MOT+DSF & w/o DSF & +App  &  \textbf{Ours} \\ 
        \midrule
        \multirow{4}{*}{KITTI-1.0s} & ADE$\downarrow$ & 0.663 & 0.582 & 0.692 & \textbf{0.471} \\ 
                                    & FDE$\downarrow$ & 1.121 & 0.978 & 1.085 & \textbf{0.763} \\ 
                                    & ASD$\uparrow$   & 1.796 & 1.730 & 1.683 & \textbf{2.351} \\ 
                                    & FSD$\uparrow$   & 3.168 & 3.052 & 3.194 & \textbf{4.071} \\ 
        \multirow{4}{*}{KITTI-3.0s} & ADE$\downarrow$ & 1.729 & 1.564 & 2.104 & \textbf{1.319} \\
                                    & FDE$\downarrow$ & 3.086 & 2.893 & 3.536 & \textbf{2.299} \\
                                    & ASD$\uparrow$   & 3.196 & 3.416 & 2.951 & \textbf{5.843} \\
                                    & FSD$\uparrow$   & 5.776 & 6.168 & 5.895 & \textbf{10.123} \\ 
        \bottomrule
        \end{tabular}}
    \end{minipage}
    \begin{minipage}{0.46\linewidth}
    	\centering
    	\vspace{0.5cm}
        \includegraphics[trim=0.3cm 0cm 0.3cm 0.3cm, clip=true, width=0.95\linewidth]{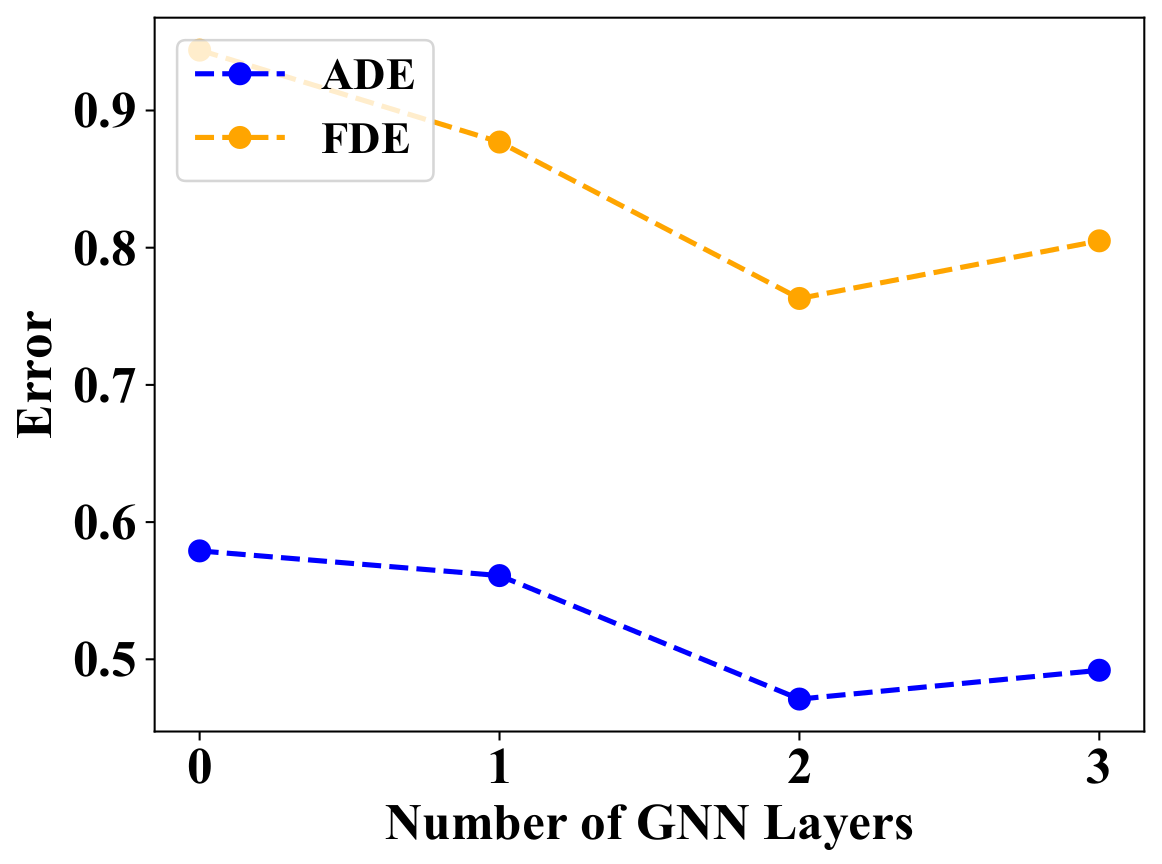}
        \vspace{-0.05cm}
        \label{fig:trajectory_gnn}
    \end{minipage}
    \vspace{-0.7cm}
\end{table}

\vspace{-0.1cm}
\subsection{Evaluating Trajectory Forecasting}
\vspace{-0.05cm}

\noindent\textbf{Evaluation Metrics.} We use two standard metrics: Average Displacement Error (ADE) \cite{alahi2016} and Final Displacement Error (FDE) for accuracy evaluation. To evaluate diversity of the trajectory samples and penalize similar samples, we use the Average Self Distance (ASD) and Final Self Distance (FSD) metrics proposed in \cite{yuan2019diverse} for sample diversity evaluation. 

\vspace{2mm}\noindent\textbf{Baselines.} We compare against S.O.T.A. methods designed for vehicle trajectory forecasting \cite{Deo2018,Chandra2019,Chandra2019_2} and person trajectory forecasting \cite{Gupta2018}, among which \cite{Chandra2019_2} also leverages GNNs. As all baselines and our method are probabilistic approaches, we follow \cite{Gupta2018} and use $20$ samples for all methods during evaluation. Following standard trajectory forecasting evaluation, which uses GT past trajectories (not estimated trajectories by a MOT module) to predict future trajectories, we feed GT trajectories in frames $t\in \{-H, \ldots, 0\}$ to baselines and predict trajectories in frames $t\in \{1, \ldots, T\}$. To have a relatively fair comparison to baselines, our method takes GT past trajectories in frames $t\in \{-H, \ldots, -1\}$ and GT detections in frame $0$ as inputs, and perform MOT in the current frame $0$ and predict trajectories in frames $t\in \{1, \ldots, T\}$ in parallel. As a result, the comparison is relatively fair as both baselines and our method have access to GT object information up to frame $0$, though our method does not use GT object identity in the frame $0$ (our method only uses GT detections in the frame $0$ but does not know which past trajectory each detection is associated to). Note that, one can also view that the baseline methods perform sequential 3D MOT and trajectory forecasting (as shown in Fig.~\ref{fig:teaser} top) but with a perfect object association in the frame $0$ while our method does MOT and forecasting in parallel.

\vspace{2mm}\noindent\textbf{Results.} We summarize trajectory forecasting results on the KITTI and nuScenes datasets in Table \ref{tab:trajectory_quan}. Our trajectory forecasting module, which (1) is jointly trained with a 3D MOT head in parallel, (2) uses GNNs for feature interaction and (3) uses diversity sampling, outperforms the baselines in both accuracy and diversity metrics. These results validate the advantage of our novel parallelized framework over prior cascaded track-forecast baselines (even using perfect tracking results in the frame $0$ from GT, let alone using real-world 3D MOT modules with imperfect results). Also, our method outperforms baselines by a large margin in the long-horizon (i.e., 3.0s) prediction setting. This is because our method has a higher sample efficiency and can cover different modes of the future trajectory distribution. We show qualitative results of our method on the KITTI dataset in Fig. \ref{fig:trajectory_qua} with plausible and diverse trajectory predictions drawn on the ground.

\vspace{2mm}\noindent\textbf{Ablation Study.} We first verify if our parallelized joint tracking/forecasting optimization and diversity sampling function improves performance of the trajectory forecasting module on the KITTI dataset. In Table \ref{tab:trajectory_ablation}, we denote our model trained without the MOT head and DSF as w/o MOT+DSF, i.e., a standard multi-modal trajectory forecasting model based on GNNs and CVAE. Then, we add one module at a time. We first add the MOT head back, denoted as w/o DSF in Table \ref{tab:trajectory_ablation}, showing clear improvement in accuracy metrics and slight improvement in diversity metrics. We believe it is because that the auxiliary 3D MOT objective improves the shared feature learning, which is helpful to trajectory forecasting. Moreover, after adding the DSF module, denoted as \textbf{Ours} in Table \ref{tab:trajectory_ablation}, we see further improvement in both accuracy and diversity metrics. Also, another interesting ablation is to see if only using motion feature is the best option for our PTP framework compared to using both motion and appearance features as in \cite{Weng2020_GNN3DMOT}. Our results (+App) show that adding appearance leads to lower performance on ADE/FDE compared to w/o DSF, which we believe is because appearance feature is not meaningful to trajectory forecasting. In the figure next to Table \ref{tab:trajectory_ablation}, we also show the effect of GNN layers on forecasting performance. We observed that ADE/FDE are decreased when using GNNs (\emph{e.g.}, 2 layers) compared to not using GNNs (\emph{i.e.}, 0 layer), showing the effectiveness of GNN-based feature interaction for trajectory forecasting in our proposed parallelized framework. 


\section{Conclusion}

We proposed a \textbf{P}arallelized 3D \textbf{T}racking and \textbf{P}rediction (PTP) framework that can avoid compounding errors and showed that it is beneficial to achieve both tasks under one unified framework through shared feature learning and joint optimization. Also, we incorporated two novel computational units into our approach: (1)~a GNN-based feature interaction, which is introduced for the first time to a joint tracking and prediction framework; (2)~a diversity sampling technique that improves sample efficiency for multi-agent trajectory forecasting. Through experiments, we established new state-of-the-art performance on both 3D MOT and trajectory forecasting, showing that the proposed method is effective.

\vspace{-0.1cm}

\bibliographystyle{IEEEtran}
\bibliography{IEEEabrv,main}

\begin{thebibliography}{10}
\providecommand{\url}[1]{#1}
\csname url@rmstyle\endcsname
\providecommand{\newblock}{\relax}
\providecommand{\bibinfo}[2]{#2}
\providecommand\BIBentrySTDinterwordspacing{\spaceskip=0pt\relax}
\providecommand\BIBentryALTinterwordstretchfactor{4}
\providecommand\BIBentryALTinterwordspacing{\spaceskip=\fontdimen2\font plus
\BIBentryALTinterwordstretchfactor\fontdimen3\font minus
  \fontdimen4\font\relax}
\providecommand\BIBforeignlanguage[2]{{%
\expandafter\ifx\csname l@#1\endcsname\relax
\typeout{** WARNING: IEEEtran.bst: No hyphenation pattern has been}%
\typeout{** loaded for the language `#1'. Using the pattern for}%
\typeout{** the default language instead.}%
\else
\language=\csname l@#1\endcsname
\fi
#2}}

\bibitem{Wang2018}
S.~Wang, D.~Jia, and X.~Weng, ``{Deep Reinforcement Learning for Autonomous
  Driving},'' \emph{arXiv:1811.11329}, 2018.

\bibitem{Weng2020_SPF2}
X.~Weng, J.~Wang, S.~Levine, K.~Kitani, and N.~Rhinehart, ``{Inverting the Pose
  Forecasting Pipeline with SPF2: Sequential Pointcloud Forecasting for
  Sequential Pose Forecasting},'' \emph{CoRL}, 2020.

\bibitem{Weng2020_AB3DMOT}
X.~Weng, J.~Wang, D.~Held, and K.~Kitani, ``{3D Multi-Object Tracking: A
  Baseline and New Evaluation Metrics},'' \emph{IROS}, 2020.

\bibitem{Zhang2019}
W.~Zhang, H.~Zhou, S.~Sun, Z.~Wang, J.~Shi, and C.~C. Loy, ``{Robust
  Multi-Modality Multi-Object Tracking},'' \emph{ICCV}, 2019.

\bibitem{Frossard2018}
D.~Frossard and R.~Urtasun, ``{End-to-End Learning of Multi-Sensor 3D Tracking
  by Detection},'' \emph{ICRA}, 2018.

\bibitem{alahi2016}
A.~{Alahi}, K.~{Goel}, V.~{Ramanathan}, A.~{Robicquet}, L.~{Fei-Fei}, and
  S.~{Savarese}, ``{Social LSTM: Human Trajectory Prediction in Crowded
  Spaces},'' \emph{CVPR}, 2016.

\bibitem{Gupta2018}
A.~Gupta, J.~Johnson, L.~Fei-Fei, S.~Savarese, and A.~Alahi, ``{Social GAN:
  Socially Acceptable Trajectories with Generative Adversarial Networks},''
  \emph{CVPR}, 2018.

\bibitem{Kosaraju2019}
V.~Kosaraju, A.~Sadeghian, R.~Mart{\'{i}}n-Mart{\'{i}}n, I.~Reid, S.~H.
  Rezatofighi, and S.~Savarese, ``{Social-BiGAT: Multimodal Trajectory
  Forecasting using Bicycle-GAN and Graph Attention Networks},''
  \emph{NeurIPS}, 2019.

\bibitem{Ivanovic2019}
B.~Ivanovic and M.~Pavone, ``{The Trajectron: Probabilistic Multi-Agent
  Trajectory Modeling With Dynamic Spatiotemporal Graphs},'' \emph{ICCV}, 2019.

\bibitem{Lee2017}
N.~Lee, W.~Choi, P.~Vernaza, C.~B. Choy, P.~H. Torr, and M.~Chandraker,
  ``{DESIRE: Distant Future Prediction in Dynamic Scenes with Interacting
  Agents},'' \emph{CVPR}, 2017.

\bibitem{Chandra2019}
R.~Chandra, U.~Bhattacharya, A.~Bera, and D.~Manocha, ``{TraPHic: Trajectory
  Prediction in Dense and Heterogeneous Traffic Using Weighted Interactions},''
  \emph{CVPR}, 2019.

\bibitem{Deo2018}
N.~Deo and M.~M. Trivedi, ``{Convolutional Social Pooling for Vehicle
  Trajectory Prediction},'' \emph{CVPRW}, 2018.

\bibitem{Yuan2021_AgentFormer}
Y.~Yuan, X.~Weng, Y.~Ou, and K.~Kitani, ``{AgentFormer: Agent-Aware
  Transformers for Socio-Temporal Multi-Agent Forecasting},''
  \emph{arXiv:2103.14023}, 2021.

\bibitem{Li2019}
X.~Li, X.~Ying, and M.~C. Chuah, ``{GRIP: Graph-Based Interaction-Aware
  Trajectory Prediction},'' \emph{ITSC}, 2019.

\bibitem{Chandra2019_2}
R.~Chandra, T.~Guan, S.~Panuganti, T.~Mittal, U.~Bhattacharya, A.~Bera, and
  D.~Manocha, ``{Forecasting Trajectory and Behavior of Road-Agents Using
  Spectral Clustering in Graph-LSTMs},'' \emph{arXiv:1912.01118}, 2019.

\bibitem{kulesza2012determinantal}
A.~Kulesza, B.~Taskar, \emph{et~al.}, ``{Determinantal Point Processes for
  Machine Learning},'' \emph{Foundations and Trends in Machine Learning}, 2012.

\bibitem{Baser2019}
E.~Baser, V.~Balasubramanian, P.~Bhattacharyya, and K.~Czarnecki, ``{FANTrack:
  3D Multi-Object Tracking with Feature Association Network},'' \emph{IV},
  2020.

\bibitem{Hu2019}
H.-N. Hu, Q.-Z. Cai, D.~Wang, J.~Lin, M.~Sun, P.~Kr{\"{a}}henb{\"{u}}hl,
  T.~Darrell, and F.~Yu, ``{Joint Monocular 3D Vehicle Detection and
  Tracking},'' \emph{ICCV}, 2019.

\bibitem{Cherabier2017}
I.~Cherabier, C.~Hane, M.~R. Oswald, and M.~Pollefeys, ``{PointNet: Deep
  Learning on Point Sets for 3D Classification and Segmentation},''
  \emph{CVPR}, 2017.

\bibitem{Simon2019}
M.~Simon, K.~Amende, A.~Kraus, J.~Honer, T.~S{\"{a}}mann, H.~Kaulbersch,
  S.~Milz, and H.~M. Gross, ``{Complexer-YOLO: Real-Time 3D Object Detection
  and Tracking on Semantic Point Clouds},'' \emph{CVPRW}, 2019.

\bibitem{Weng2020_GNN3DMOT}
X.~Weng, Y.~Wang, Y.~Man, and K.~Kitani, ``{GNN3DMOT: Graph Neural Network for
  3D Multi-Object Tracking with 2D-3D Multi-Feature Learning},'' \emph{CVPR},
  2020.

\bibitem{Kitani2012}
K.~M. Kitani, B.~D. Ziebart, J.~A. Bagnell, and M.~Hebert, ``{Activity
  Forecasting},'' \emph{ECCV}, 2012.

\bibitem{robicquet2016}
A.~Robicquet, A.~Sadeghian, A.~Alahi, and S.~Savarese, ``{Learning Social
  Etiquette: Human Trajectory Understanding In Crowded Scenes},'' \emph{ECCV},
  2016.

\bibitem{yuan2019ego}
Y.~Yuan and K.~Kitani, ``{Ego-Pose Estimation and Forecasting as Real-Time PD
  Control},'' \emph{ICCV}, 2019.

\bibitem{Rhinehart2019}
N.~Rhinehart, R.~McAllister, K.~Kitani, and S.~Levine, ``{PRECOG: PREdiction
  Conditioned On Goals in Visual Multi-Agent Settings},'' \emph{ICCV}, 2019.

\bibitem{Rhinehart2018}
N.~Rhinehart, M.~Kris, and P.~Vernaza, ``{R2P2: A ReparameteRized Pushforward
  Policy for Diverse, Precise Generative Path Forecasting},'' \emph{ECCV},
  2018.

\bibitem{Wang2020_GNNDetTrk}
Y.~Wang, K.~Kitani, and X.~Weng, ``{Joint Object Detection and Multi-Object
  Tracking with Graph Neural Networks},'' \emph{ICRA}, 2021.

\bibitem{Zeng2019}
W.~Zeng, W.~Luo, S.~Suo, A.~Sadat, B.~Yang, S.~Casas, and R.~Urtasun,
  ``{End-to-End Interpretable Neural Motion Planner},'' \emph{CVPR}, 2019.

\bibitem{Casas2018}
S.~Casas, W.~Luo, and R.~Urtasun, ``{IntentNet: Learning to Predict Intention
  from Raw Sensor Data},'' \emph{CoRL}, 2018.

\bibitem{Luo2018}
W.~Luo, B.~Yang, and R.~Urtasun, ``{Fast and Furious: Real Time End-to-End 3D
  Detection, Tracking and Motion Forecasting with a Single Convolutional
  Net},'' \emph{CVPR}, 2018.

\bibitem{Liang2020}
M.~Liang, B.~Yang, W.~Zeng, Y.~Chen, R.~Hu, S.~Casas, and R.~Urtasun,
  ``{PnPNet: End-to-End Perception and Prediction with Tracking in the Loop},''
  \emph{CVPR}, 2020.

\bibitem{Gori2005}
M.~Gori, G.~Monfardini, and F.~Scarselli, ``{A New Model for Learning in Graph
  Domains},'' \emph{IJCNN}, 2005.

\bibitem{Chen2019}
Y.~Chen, M.~Rohrbach, Z.~Yan, S.~Yan, J.~Feng, and Y.~Kalantidis,
  ``{Graph-Based Global Reasoning Networks},'' \emph{CVPR}, 2019.

\bibitem{Zhang2019_graph}
L.~Zhang, X.~Li, A.~Arnab, K.~Yang, Y.~Tong, and P.~H.~S. Torr, ``{Dual Graph
  Convolutional Network for Semantic Segmentation},'' \emph{BMVC}, 2019.

\bibitem{Wang2018_2}
X.~Wang and A.~Gupta, ``{Videos as Space-Time Region Graphs},'' \emph{ECCV},
  2018.

\bibitem{Li2019_action}
M.~Li, S.~Chen, X.~Chen, Y.~Zhang, Y.~Wang, and Q.~Tian, ``{Actional-Structural
  Graph Convolutional Networks for Skeleton-based Action Recognition},''
  \emph{CVPR}, 2019.

\bibitem{Cheng2019}
L.~Shi, Y.~Zhang, J.~Cheng, and H.~Lu, ``{Skeleton-Based Action Recognition
  with Directed Graph Neural Networks},'' \emph{CVPR}, 2019.

\bibitem{Zhao2019}
R.~Zhao, K.~Wang, H.~Su, and Q.~Ji, ``{Bayesian Graph Convolution LSTM for
  Skeleton Based Action Recognition},'' \emph{ICCV}, 2019.

\bibitem{Gao2019}
J.~Gao, T.~Zhang, and C.~Xu, ``{Graph Convolutional Tracking},'' \emph{CVPR},
  2019.

\bibitem{seroussi1994algorithm}
B.~Seroussi and J.-L. Golmard, ``{An Algorithm Directly Finding the K Most
  Probable Configurations in Bayesian Networks},'' \emph{International Journal
  of Approximate Reasoning}, 1994.

\bibitem{batra2012diverse}
D.~Batra, P.~Yadollahpour, A.~Guzman-Rivera, and G.~Shakhnarovich, ``{Diverse
  m-Best Solutions in Markov Random Fields},'' \emph{ECCV}, 2012.

\bibitem{lee2016stochastic}
S.~Lee, S.~P.~S. Prakash, M.~Cogswell, V.~Ranjan, D.~Crandall, and D.~Batra,
  ``{Stochastic Multiple Choice Learning for Training Diverse Deep
  Ensembles},'' \emph{NIPS}, 2016.

\bibitem{hsiao2018creating}
W.-L. Hsiao and K.~Grauman, ``{Creating Capsule Wardrobes from Fashion
  Images},'' \emph{CVPR}, 2018.

\bibitem{gong2014diverse}
B.~Gong, W.-L. Chao, K.~Grauman, and F.~Sha, ``{Diverse Sequential Subset
  Selection for Supervised Video Summarization},'' \emph{NIPS}, 2014.

\bibitem{azadi2017learning}
S.~Azadi, J.~Feng, and T.~Darrell, ``{Learning Detection with Diverse
  Proposals},'' \emph{CVPR}, 2017.

\bibitem{huang2015we}
D.-A. Huang, M.~Ma, W.-C. Ma, and K.~M. Kitani, ``{How do We Use Our Hands?
  Discovering a Diverse Set of Common Grasps},'' \emph{CVPR}, 2015.

\bibitem{che2016mode}
T.~Che, Y.~Li, A.~P. Jacob, Y.~Bengio, and W.~Li, ``{Mode Regularized
  Generative Adversarial Networks},'' \emph{ICLR}, 2017.

\bibitem{arjovsky2017wasserstein}
M.~Arjovsky, S.~Chintala, and L.~Bottou, ``{Wasserstein Generative Adversarial
  Networks},'' \emph{ICML}, 2017.

\bibitem{gulrajani2017improved}
I.~Gulrajani, F.~Ahmed, M.~Arjovsky, V.~Dumoulin, and A.~C. Courville,
  ``{Improved Training of Wasserstein GANs},'' \emph{NIPS}, 2017.

\bibitem{elfeki2018gdpp}
M.~Elfeki, C.~Couprie, M.~Riviere, and M.~Elhoseiny, ``{GDPP: Learning Diverse
  Generations Using Determinantal Point Process},'' \emph{ICML}, 2019.

\bibitem{zhao2017infovae}
S.~Zhao, J.~Song, and S.~Ermon, ``{InfoVAE: Balancing Learning and Inference in
  Variational Autoencoders},'' \emph{AAAI}, 2019.

\bibitem{tolstikhin2017wasserstein}
I.~Tolstikhin, O.~Bousquet, S.~Gelly, and B.~Schoelkopf, ``{Wasserstein
  Auto-Encoders},'' \emph{ICLR}, 2018.

\bibitem{he2019lagging}
J.~He, D.~Spokoyny, G.~Neubig, and T.~Berg-Kirkpatrick, ``{Lagging Inference
  Networks and Posterior Collapse in Variational Autoencoders},'' \emph{ICLR},
  2019.

\bibitem{yuan2019diverse}
Y.~Yuan and K.~Kitani, ``{Diverse Trajectory Forecasting with Determinantal
  Point Processes},'' \emph{ICLR}, 2020.

\bibitem{yuan2020dlow}
Y.~Yuan and K.~M. Kitani, ``{Dlow: Diversifying Latent Flows for Diverse Human
  Motion Prediction},'' \emph{ECCV}, 2020.

\bibitem{Morris2019}
C.~Morris, M.~Ritzert, M.~Fey, W.~L. Hamilton, J.~E. Lenssen, G.~Rattan, and
  M.~Grohe, ``{Weisfeiler and Leman Go Neural: Higher-Order Graph Neural
  Networks},'' \emph{AAAI}, 2019.

\bibitem{WKuhn1955}
H.~{W Kuhn}, ``{The Hungarian Method for the Assignment Problem},'' \emph{Naval
  Research Logistics Quarterly}, 1955.

\bibitem{Geiger2012}
A.~Geiger, P.~Lenz, and R.~Urtasun, ``{Are We Ready for Autonomous Driving? the
  KITTI Vision Benchmark Suite},'' \emph{CVPR}, 2012.

\bibitem{Caesar2019}
H.~Caesar, V.~Bankiti, A.~H. Lang, S.~Vora, V.~E. Liong, and Q.~Xu,
  ``{nuScenes: A Multimodal Dataset for Autonomous Driving},'' \emph{CVPR},
  2020.

\bibitem{Milan2016}
A.~Milan, L.~Leal-Taixe, I.~Reid, S.~Roth, and K.~Schindler, ``{MOT16: A
  Benchmark for Multi-Object Tracking},'' \emph{TPAMI}, 2016.

\bibitem{Shi2019}
S.~Shi, X.~Wang, and H.~Li, ``{PointRCNN: 3D Object Proposal Generation and
  Detection from Point Cloud},'' \emph{CVPR}, 2019.

\bibitem{Zhu2019}
B.~Zhu, Z.~Jiang, X.~Zhou, Z.~Li, and G.~Yu, ``{Class-Balanced Grouping and
  Sampling for Point Cloud 3D Object Detection},'' \emph{CVPR}, 2019.

\end{thebibliography}

\end{document}